\documentclass[letterpaper]{article}
\usepackage[preprint]{aaai2027}
\usepackage[hyphens]{url}
\usepackage{graphicx}
\usepackage{natbib}
\usepackage{caption}
\usepackage[T1]{fontenc}
\usepackage[utf8]{inputenc}

\usepackage{algorithm}
\usepackage{algpseudocode}

\usepackage{booktabs}

\usepackage{colortbl}

\newcolumntype{G}{!{\color{black!45}\vrule width 0.35pt}}
\newcommand{\NAcell}{\textcolor{gray}{N/A}}

\usepackage[most]{tcolorbox}

\usepackage{amsmath}
\usepackage{amssymb}
\usepackage{xcolor}
\usepackage{subcaption}

\graphicspath{{Figures/}}

\algrenewcommand\algorithmicrequire{\textbf{Input:}}
\algrenewcommand\algorithmicensure{\textbf{Output:}}

\title{MOSAIC: Adversarial Co-evolution of Specialist Heuristics\\and Problem Instances for LLM-based Automated Heuristic Design}
\author{
    Oguzhan Gungordu,
    Siheng Xiong,
    Faramarz Fekri
}
\affiliations{
    Georgia Institute of Technology\\
    \{ogungordu3, sxiong45\}@gatech.edu, faramarz.fekri@ece.gatech.edu
}
\copyrighttext{Preprint. Under review.}

\begin{document}
\maketitle
\begin{abstract}
Automated heuristic design (AHD) with large language models (LLMs) has produced strong heuristics for combinatorial optimization problems (COPs). Yet existing frameworks optimize for average performance on a small fixed dataset and steer the search with ``verbal gradients'' distilled from scalar better/worse feedback. No single heuristic dominates across instance distributions, and scalar feedback tells the LLM \emph{whether} a heuristic improved, but not \emph{where} in the instance space or \emph{why}. We propose \textbf{MOSAIC}, a grid-based framework that adversarially co-evolves problem instances and specialist heuristics inside a Quality--Diversity (QD) archive indexed by structural instance features. Instances evolve to expose weaknesses of the current heuristics, and heuristics evolve to eliminate them by specializing to the newly exposed regions. Each archive cell keeps a specialist heuristic, representative instances, and insights explaining what works in its region, forming a persistent memory that accumulates over the evolutionary search. For each heuristic pair sampled from distant grid regions, an LLM-guided evolutionary loop generates discriminative instances, and a decision tree identifies the feature-space regions where each heuristic wins. A reflection LLM then contrasts the two heuristics to produce \emph{multi-directional} insights that persist in those regions and guide crossover and mutation. The archive is simultaneously a co-evolved benchmark of discriminative instances and a pool of region specialist heuristics, from which greedy selection extracts a compact complementary portfolio. Across COPs, test sizes, and LLM backbones, the portfolio consistently outperforms state-of-the-art LLM-based AHD methods, and the co-evolved instances attain higher feature-space coverage and stronger heuristic discrimination than evolutionary instance-generation baselines.
\end{abstract}

\section{Introduction}
\label{sec:intro}

COPs in routing, scheduling, and logistics are predominantly solved with heuristics, as NP-hardness makes exact solution impractical at scale~\citep{Ausiello2012}. Designing heuristics by hand requires substantial domain knowledge and rarely transfers across problem classes. This motivates AHD, traditionally realized through hyper-heuristics and genetic programming~\citep{burke2019hyper,branke2016}.

LLMs have opened a new direction for AHD by combining code generation with algorithmic reasoning~\citep{liu2023algorithm}. FunSearch~\citep{funsearch} and EoH~\citep{eoh} evolve executable heuristics with LLM-based variation operators; ReEvo~\citep{ye2024reevo} guides the search with verbal reflections on past heuristics; HSEvo~\citep{hsevo} promotes diversity via harmony search; MCTS-AHD~\citep{zheng2025montecarlotreesearch} explores the heuristic space with tree search; PathWise~\citep{gungordu2026pathwise} plans evolutionary actions through state-aware multi-agent reasoning; and EoH-S~\citep{eohs} extends the output from a single heuristic to a small complementary set.

These methods optimize for average performance on a small fixed training set and steer the search with scalar better/worse feedback. \textbf{Specialist heuristics:} the No Free Lunch (NFL) theorem~\citep{wolpert1997nfl} suggests that no single heuristic dominates across instance distributions, and LLM-evolved heuristics generalize poorly beyond their training distribution~\citep{sim2025hypebenchmarkingllmevolvedheuristics}. Even set-based extensions specialize only to the regimes their fixed dataset contains. \textbf{Discriminative instances:} specializing a heuristic requires knowing \emph{where} and \emph{why} heuristics differ, yet scalar feedback carries no directional information~\citep{nie2024importancedirectionalfeedbackllmbased} and each reflection is discarded after a single iteration. Instance Space Analysis (ISA)~\citep{SMITHMILES201412instanceinisaspace} maps the feature-space regions where algorithms win, and evolutionary instance generation~\citep{SMITHMILES2015evolveinstance,bossek2022gecco} evolves diverse, discriminating instances, but both operate post hoc on frozen solvers.

We propose \textbf{M}ap \textbf{O}f \textbf{S}pecialists via \textbf{A}dversarial \textbf{I}nstance \textbf{C}o-evolution (MOSAIC), a grid-based co-evolutionary framework that couples these two requirements adversarially: instance generation exposes weaknesses of current heuristics by evolving hard, discriminative instances, and heuristic generation eliminates them by specializing to newly exposed regions. Both populations share a QD archive~\citep{mouret2015mapelites} indexed by bounded structural instance features. Each iteration contrasts heuristic pairs from distant regions: an LLM-guided evolutionary loop evolves instances that maximally separate the pair, a decision tree maps the feature-space regions where each heuristic wins, and a contrastive reflection distills \emph{multi-directional} insights that persist in those regions and steer crossover and region-aware mutation. The resulting archive is simultaneously a co-evolved benchmark of discriminative instances and a pool of specialist heuristics adapted to different regions.

Our contributions are as follows. \textbf{(1)} We introduce the first grid-based AHD framework that co-evolves instances and heuristics under an instance-space view, coupling feature-space coverage with algorithmic discrimination in an adversarial objective. \textbf{(2)} We replace scalar feedback with \emph{contrastive multi-directional reflection}, whose insights are anchored by a decision tree to the feature-space regions where each parent wins. \textbf{(3)} We use the QD archive as a persistent memory of region specialists, representative instances, and localized insights, accumulating cross-regional knowledge across iterations. \textbf{(4)} Experiments across COPs show that \textbf{(a)}~MOSAIC consistently outperforms state-of-the-art LLM-based AHD methods, and \textbf{(b)}~its co-evolved instances, produced by an LLM-guided evolutionary generator, attain higher feature-space coverage and stronger heuristic discrimination than evolutionary instance-generation baselines.

\begin{figure*}[t]
    \centering
    \includegraphics[width=\linewidth]{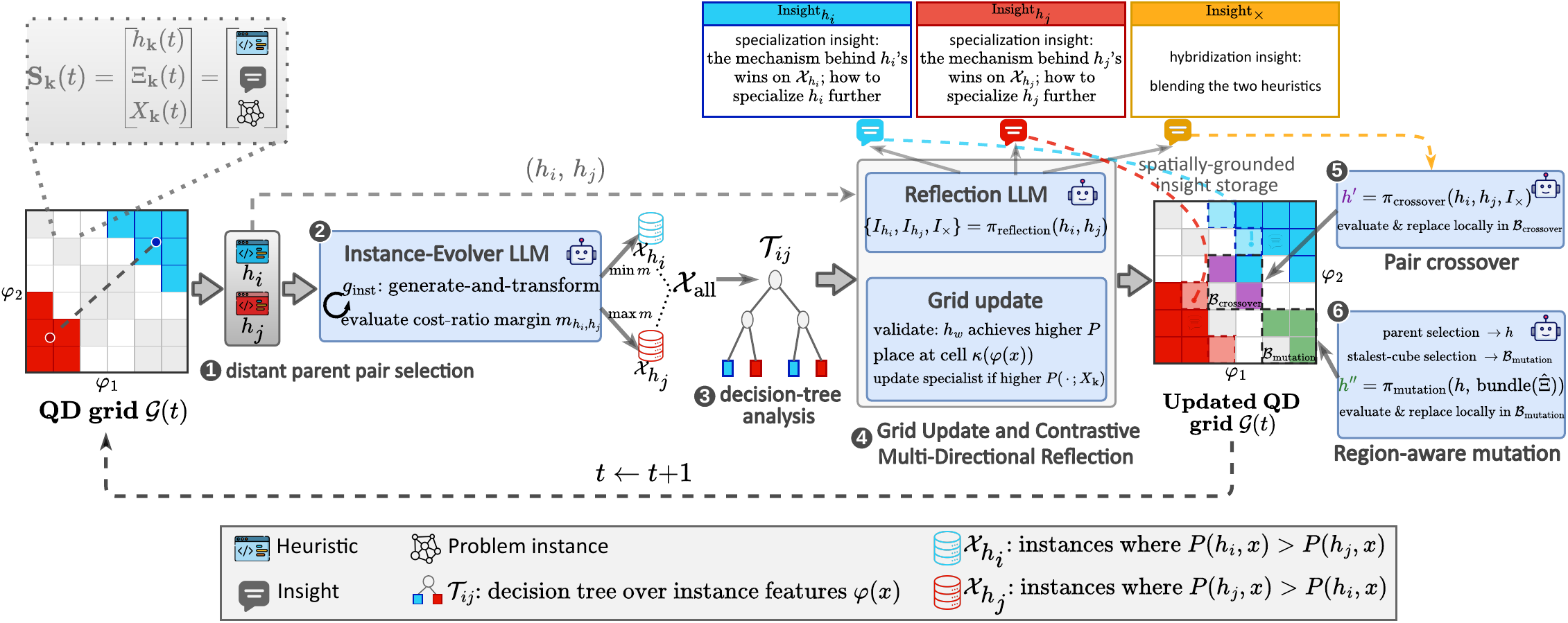}
    \caption{Overview of MOSAIC on a QD grid archive (axes: instance features); each cell stores a specialist heuristic, localized insights, and representative instances. \textbf{(1)}~A parent pair $(h_i, h_j)$ is sampled from distant grid regions. \textbf{(2)}~$\boldsymbol{\pi}_{\text{instance-evolver}}$ evolves discriminative instance sets $\mathcal{X}_{h_i}$ and $\mathcal{X}_{h_j}$. \textbf{(3)}~A decision tree $\mathcal{T}_{ij}$ identifies each heuristic's winning feature-space regions. \textbf{(4)}~Validated instances and their winners update the grid, and $\boldsymbol{\pi}_{\text{reflection}}$ produces the multi-directional insights $\{I_{h_i}, I_{h_j}, I_{\times}\}$, stored in those regions. \textbf{(5)}~$\boldsymbol{\pi}_{\text{crossover}}$ blends the parents guided by $I_{\times}$, and \textbf{(6)}~$\boldsymbol{\pi}_{\text{mutation}}$ specializes a parent toward a stale region using its stored insights; offspring replace incumbent specialists in their evaluation cubes.}
    \label{fig:overview}
\end{figure*}

\section{Preliminaries}
\label{sec:prelim}

We formalize the AHD problem, the instance feature space, the QD grid archive, and the co-evolutionary objective, which together form the basis of the framework.

\subsection{Combinatorial Optimization and Automated Heuristic Design}
\label{sec:prelim_cop}

AHD targets a COP with instance space $\mathcal{X}$ and solution space $\mathcal{S}$. A \emph{heuristic} is a program $h \in \mathcal{H}$ mapping each instance to a feasible solution, $h: \mathcal{X} \to \mathcal{S}$, evaluated by a cost function $f: \mathcal{S} \to \mathbb{R}$. Performance over a dataset $\mathcal{D}$ is
\begin{equation}
    P(h; \mathcal{D}) = \mathbb{E}_{x \sim \mathcal{D}}\big[-f(h(x))\big],
    \label{eq:performance}
\end{equation}
with per-instance performance $P(h, x) = -f(h(x))$. Under an evaluation budget, traditional AHD seeks $h^* = \arg\max_{h \in \mathcal{H}} P(h; \mathcal{D})$.

\subsection{Instance Feature Space}
\label{sec:prelim_features}

Each instance $x$ is summarized by a $d$-dimensional feature vector $\boldsymbol{\varphi}(x) = (\varphi_1(x), \ldots, \varphi_d(x)) \in \mathbb{R}^d$, where each $\varphi_i$ extracts a structural property of $x$ (e.g., distance-distribution and nearest-neighbor-graph statistics), as in ISA~\citep{SMITHMILES2012875}. Each feature has a known type $\tau_i \in \{\texttt{integer}, \texttt{continuous}\}$ and a bounded range $[\ell_i, u_i]$ derived analytically from the feature's definition and the instance size $n$~\citep{HEINS2023123}. The feature vector places each instance in the grid archive of Section~\ref{sec:prelim_grid}, whose axes and per-axis resolution are fixed by these types and bounds.

\subsection{Quality--Diversity Grid Archive}
\label{sec:prelim_grid}

The central data structure of our framework is a QD grid archive $\mathcal{G}$. It is a $d$-dimensional grid discretizing the feature space, one axis per feature. Each cell $\mathbf{c}_{\mathbf{k}} \in \mathcal{G}$ at coordinate $\mathbf{k} \in \mathbb{Z}^d$ stores a state vector that evolves over the iterations $t$ of the co-evolutionary loop,
\begin{equation}
    \mathbf{S}_{\mathbf{k}}(t) = \big[\, h_{\mathbf{k}}(t),\;\Xi_{\mathbf{k}}(t),\; X_{\mathbf{k}}(t)\,\big]^\top,
\end{equation}
consisting of:
\begin{itemize}
    \item a \emph{specialist} $h_{\mathbf{k}} \in \mathcal{H}$, a region-adapted heuristic maintained for instances whose features fall in this cell;
    \item a set of localized insights $\Xi_{\mathbf{k}}$, natural-language reflections explaining what works in this region;
    \item a set $X_{\mathbf{k}} \subseteq \mathcal{X}$ of up to $M{=}3$ representative instances.
\end{itemize}
Each cell maintains an update counter $\nu_{\mathbf{k}}$ recording how many times its specialist has been replaced. We denote the set of \emph{filled} cells $\mathcal{C}(t) = \{\mathbf{k} : |X_{\mathbf{k}}(t)| > 0\}$.

\paragraph{Coordinate mapping.}
Each instance is assigned to the cell $\mathbf{k}(x) = \kappa(\boldsymbol{\varphi}(x))$, where the map $\kappa: \mathbb{R}^d \to \mathbb{Z}^d$ acts coordinate-wise, its $i$-th coordinate $k_i$ discretizing feature $\varphi_i$ according to its type: an integer feature gets one cell per value, and a continuous feature is split into $R_i{=}10$ equal-width bins,
\begin{equation}
    k_i = \begin{cases}
        \lfloor \varphi_i(x) \rceil - \ell_i & \tau_i = \texttt{integer}, \\[4pt]
        \left\lfloor \dfrac{\varphi_i(x) - \ell_i}{(u_i - \ell_i)/R_i} \right\rfloor & \tau_i = \texttt{continuous},
    \end{cases}
    \label{eq:discretization}
\end{equation}
where $\lfloor \cdot \rceil$ rounds to the nearest integer. Coordinates are clamped to each axis's range, so the per-axis resolution is $u_i - \ell_i + 1$ for integer and $R_i$ for continuous features. The average performance of $h$ in cell $\mathbf{k}$ is $P(h; X_{\mathbf{k}}) = \frac{1}{|X_{\mathbf{k}}|} \sum_{x \in X_{\mathbf{k}}} P(h, x)$, and a candidate replaces the incumbent specialist if it achieves strictly higher $P(h; X_{\mathbf{k}})$.

\paragraph{Cube neighborhood.}
A \emph{cube} $\mathcal{B}(\bar{\mathbf{k}}, \rho) \subseteq \mathcal{G}$ is an axis-aligned neighborhood centered at $\bar{\mathbf{k}}$ whose side along axis $i$ spans a fraction $\rho{=}0.4$ of that axis's resolution, expanded by adding the nearest filled cells in normalized coordinates until at least $T{=}64$ cells are reached. Cubes are the local evaluation contexts for newly generated heuristics, large enough to be statistically meaningful and small enough to limit heuristic-evaluation cost to the relevant region.

\subsection{Co-Evolutionary Optimization Objective}
\label{sec:prelim_objective}

\paragraph{Why co-evolution.}
The NFL theorem motivates building a strong heuristic portfolio from \emph{specialists} adapted to different instance-space regions. Joint evolution provides this directly: diverse instances expose where heuristics differ, and diverse heuristics make those instances meaningful as discrimination targets.

\paragraph{Heuristic-generation objective.}
Given a grid with filled cells $\mathcal{C}$, the heuristic-generation problem seeks specialists $\{h_{\mathbf{k}}\}_{\mathbf{k} \in \mathcal{C}}$ maximizing total performance across cells:
\begin{equation}
    \max_{\{h_{\mathbf{k}}\}_{\mathbf{k} \in \mathcal{C}}}\; \sum_{\mathbf{k} \in \mathcal{C}} P(h_{\mathbf{k}}; X_{\mathbf{k}}).
    \label{eq:heuristic_obj}
\end{equation}
Each cell is a separate \emph{task}: $h_{\mathbf{k}}$ is optimized only for instances in cell $\mathbf{k}$. Tasks are not predefined---they \emph{emerge} as feature-space regions become populated.

\paragraph{Instance-generation objective.}
The instance-generation problem seeks instance sets that are jointly \textbf{(i)~structurally diverse} (expanding the filled set $\mathcal{C}$, measured by the coverage ratio $|\mathcal{C}|/|\mathcal{G}|$) and \textbf{(ii)~algorithmically discriminating} (separating heuristic performance via the discrimination score
\begin{equation}
    \Delta(x) = \max_{h \in \mathcal{H}_{\mathcal{G}}} f(h(x)) - \min_{h \in \mathcal{H}_{\mathcal{G}}} f(h(x)),
    \label{eq:discrimination}
\end{equation}
where $\mathcal{H}_{\mathcal{G}} = \{h : \exists\, \mathbf{k} \text{ s.t.\ } h_{\mathbf{k}} = h\}$). A well-constructed instance set distributes dominance across the heuristic portfolio rather than letting a single heuristic win everywhere; the corresponding scalar objective combines mean discrimination with a coverage incentive (Appendix~\ref{sec:appendix_objective_detail}).

\paragraph{Co-evolutionary objective.}
The two sub-problems are coupled adversarially: instance generation \emph{exposes weaknesses} in the current heuristic portfolio, while heuristic generation \emph{eliminates them} by specializing to newly discovered regions. We express this as a minimax game,
\begin{equation}
    \max_{\{h_{\mathbf{k}}\}}\, \min_{\{X_{\mathbf{k}}\}} \sum_{\mathbf{k} \in \mathcal{C}} P(h_{\mathbf{k}}; X_{\mathbf{k}}) \;-\; \lambda\, \frac{|\mathcal{C}|}{|\mathcal{G}|},
    \label{eq:objective}
\end{equation}
with $\lambda > 0$. The negative coverage term rewards the instance adversary for expanding coverage; the outer maximization seeks specialists robust to those adversarial choices. At equilibrium, each cell holds a specialist performing well on the hardest instances in its region, and the instance set covers the feature space while maximally separating heuristic performance. We operationalize the inner minimization via cost-ratio margins between heuristic pairs (Section~\ref{sec:method_isa}), a tractable proxy for portfolio-level discrimination (Eq.~\ref{eq:discrimination}).

\section{Methodology}
\label{sec:method}

MOSAIC iterates the loop shown in Figure~\ref{fig:overview}: \textbf{(1)} distant parent pair selection, \textbf{(2)} discriminative instance generation, \textbf{(3)} decision-tree analysis, \textbf{(4)} grid update with contrastive multi-directional reflection, \textbf{(5)} pair crossover, and \textbf{(6)} region-aware mutation. After the evolutionary budget is exhausted, greedy selection extracts a compact complementary test-time portfolio. Pseudocode is in Appendix~\ref{sec:pseudocodes}, prompts in Appendix~\ref{app:prompts}.

\subsection{Initialization}
\label{sec:init}

The grid is seeded from an initial dataset $\mathcal{D}_{\text{init}}$: each instance is placed in cell $\kappa(\boldsymbol{\varphi}(x))$, with at most $M$ representatives per cell. An initial population of $N_{\text{init}}{=}20$ heuristics is generated by independently prompting the LLM with the problem description and function signature; each is evaluated on filled cells, and the best per cell becomes the initial specialist.

\subsection{Parent Pair Selection}
\label{sec:method_pairs}

At each iteration, $N_{\text{pair}}{=}10$ heuristic pairs are selected for analysis (Fig.~\ref{fig:overview}, step~1), favoring pairs from distant feature-space regions, which are most likely to exhibit complementary strengths and produce discriminative instances with less effort. For each heuristic $h$ with grid presence we compute its centroid in normalized coordinates,
\begin{equation}
    \bar{\boldsymbol{\varphi}}(h) = \frac{1}{|\mathcal{C}(h)|}\sum_{\mathbf{k} \in \mathcal{C}(h)} \tilde{\mathbf{k}},
    \label{eq:centroid}
\end{equation}
where $\mathcal{C}(h) = \{\mathbf{k} : h_{\mathbf{k}} = h\}$ and $\tilde{\mathbf{k}}$ is the cell center min--max-normalized to $[0,1]$ by the feature bounds $[\ell_i, u_i]$. Previously unanalyzed pairs are ranked by centroid distance; the $N_{\text{cand}}{=}20$ most distant pairs are retained, and $N_{\text{pair}}$ pairs are sampled without replacement under inverse-rank probability $p_r \propto (r+1)^{-1}$, rank $r=1$ being the most distant pair. The sampled pairs form the analysis set $\mathcal{Q}$.

\subsection{Discriminative Instance Generation and Decision-Tree Analysis}
\label{sec:method_isa}

For each pair $(h_i, h_j) \in \mathcal{Q}$, we generate instances that maximally discriminate between them (Fig.~\ref{fig:overview}, step~2), measured by the cost-ratio margin
\begin{equation}
    m_{h_i, h_j}(x) = \frac{f(h_i(x))}{f(h_j(x))},
    \label{eq:margin}
\end{equation}
so maximizing $m_{h_i, h_j}$ yields instances on which $h_j$ dominates and minimizing it yields the opposite. Evolution runs in the two directions sequentially: we first evolve $\mathcal{X}_{h_j} \leftarrow \arg\max_x m_{h_i, h_j}(x)$, instances hard for $h_i$ and easy for $h_j$; if no such discriminative instance is found, i.e., no evolved instance attains $m_{h_i, h_j}(x) \geq 1$, the pair is \emph{dominated}, $h_i$ is its \emph{clear winner}, and the opposite direction is skipped. Otherwise we evolve $\mathcal{X}_{h_i} \leftarrow \arg\min_x m_{h_i, h_j}(x)$, instances hard for $h_j$ and easy for $h_i$; if none is found, $h_j$ is the clear winner. A pair with discriminative instances in both directions is a \emph{discriminated pair}.

\paragraph{LLM-guided evolutionary instance generation.}
For each direction, an instance-evolver LLM $\boldsymbol{\pi}_{\text{instance-evolver}}$ is queried once for a pair-specific generate-and-transform operator,
\begin{equation}
    g_{\text{inst}} = \boldsymbol{\pi}_{\text{instance-evolver}}(h_i, h_j, \text{direction}),
    \label{eq:inst_gen_llm}
\end{equation}
an executable function with two modes: given fewer than $N_{\text{inst}}{=}50$ instances it generates new ones from them, and given a full population it applies incremental structural transformations. The operator then drives an evolutionary loop:
\begin{enumerate}
    \item \textbf{\emph{Bootstrap.}} Seed instances $X_{\text{seed}}$ are drawn from the grid cells whose specialist is the heuristic to be favored ($h_j$ when maximizing $m_{h_i, h_j}$, $h_i$ when minimizing); since typically $|X_{\text{seed}}| < N_{\text{inst}}$, the call $X \gets g_{\text{inst}}(X_{\text{seed}}, N_{\text{inst}})$ runs in generation mode and returns an instance population $X$ of size $N_{\text{inst}}$.
    \item \textbf{\emph{Evaluate \& Select.}} Every $x \in X$ is scored by $m_{h_i, h_j}(x)$ and the top-$N_{\text{inst}}$ instances are kept (largest margins when maximizing, smallest when minimizing).
    \item \textbf{\emph{Transform.}} The call $X' \gets g_{\text{inst}}(X, N_{\text{inst}})$ runs in transform mode and returns $N_{\text{inst}}$ structurally modified variants of $X$; parents and offspring are re-ranked and $X \gets$ top-$N_{\text{inst}}$ of $X \cup X'$.
    \item \textbf{\emph{Iterate.}} Repeat 2)--3) for up to $G{=}10$ generations, stopping early if the best margin stagnates for $S{=}5$ generations or crosses an early-termination threshold ($\max_{x \in X} m_{h_i, h_j}(x) \geq 2$ when maximizing); the surviving instance set $X$ is returned as $\mathcal{X}_{h_j}$ or $\mathcal{X}_{h_i}$, respectively.
\end{enumerate}
The LLM contributes semantic reasoning about the heuristic pair's weaknesses; the evolutionary loop refines the instance population toward larger margins.

\paragraph{Decision-tree analysis.}
For discriminated pairs (Fig.~\ref{fig:overview}, step~3), we pool the instances from both directions, $X_{\text{all}} = \mathcal{X}_{h_j} \cup \mathcal{X}_{h_i}$, label each with its winner, $y(x) = \mathbb{I}[P(h_j, x) > P(h_i, x)]$, and fit a shallow decision tree $\mathcal{T}_{ij}$ on $\{(\boldsymbol{\varphi}(x), y(x)) : x \in X_{\text{all}}\}$. Its leaves validate which evolved instances enter which grid cells and determine which cells receive which insights.

\subsection{Grid Update and Contrastive Multi-Directional Reflection}
\label{sec:method_grid_update}

The outcome of the pair analysis, a dominated or a discriminated pair, drives the grid update (Fig.~\ref{fig:overview}, step~4). \textbf{Dominated pair:} the clear winner takes over the loser's cells without re-evaluation, since dominance was established adversarially, and evolved instances expand the grid into new cells with the winner as specialist. \textbf{Discriminated pair:} each leaf of $\mathcal{T}_{ij}$ predicts a winning heuristic $h_w$ for its region, and its instances are processed as follows:
\begin{enumerate}
    \item \textbf{\emph{Validate.}} Keep only the instances on which $h_w$ achieves a higher $P$ value than the other parent.
    \item \textbf{\emph{Place.}} Assign each validated instance to its cell $\kappa(\boldsymbol{\varphi}(x))$: create the cell if empty and grow it to $M$ instances. Keep the $M$ where its specialist performs best.
    \item \textbf{\emph{Update specialist.}} In each cell that received a validated instance, the leaf's winning parent $h_w$ replaces the incumbent specialist if it achieves higher $P(\cdot; X_{\mathbf{k}})$.
\end{enumerate}
One mechanism thus \emph{expands}, \emph{refines}, and \emph{updates} the archive simultaneously.

\paragraph{Contrastive multi-directional reflection.}
For each discriminated pair, we prompt a reflection LLM $\boldsymbol{\pi}_{\text{reflection}}$ with the two parent heuristics,
\begin{equation}
    \{I_{h_i}, I_{h_j}, I_{\times}\} = \boldsymbol{\pi}_{\text{reflection}}(h_i, h_j),
    \label{eq:reflection}
\end{equation}
returning three structured insights:
\begin{itemize}
    \item $I_{h_i}$ is a specialization insight distilling the algorithmic mechanism behind $h_i$'s wins on $\mathcal{X}_{h_i}$ and how to specialize $h_i$ further on such instances;
    \item $I_{h_j}$ does the same for $h_j$ on $\mathcal{X}_{h_j}$;
    \item $I_{\times}$ is a hybridization insight blending the two heuristics.
\end{itemize}
Contrastive reasoning over heuristics winning in different regions converts a pairwise comparison into \emph{multi-directional} feedback: two region-specific specialization directions and one hybridization direction. For dominated pairs, a global reflection contrasting the clear winner and loser replaces the three insights; it is stored in the winner's cells and the loser's former cells, and takes the role of $I_{\times}$ in crossover.

\paragraph{Spatially-grounded insight storage.}
Each insight is stored in every cell that received a validated instance from its leaf, tagged with the iteration $t$ at which it was generated. The cell-wise set $\Xi_{\mathbf{k}}$ accumulates across iterations and \emph{persists beyond the current specialist} $h_{\mathbf{k}}$, forming a durable memory of region $\mathbf{k}$. From a Bayesian viewpoint, $\Xi_{\mathbf{k}}$ acts as a region-aware prior (Appendix~\ref{sec:appendix_bayesian}).

\subsection{LLM-Guided Insight-Driven Heuristic Generation}
\label{sec:method_operators}

Given the updated grid, each iteration generates heuristics through two LLM-driven operators that use the stored insights: crossover blends the parents of each analyzed pair, and mutation adapts a single parent toward a chosen grid region, each operator producing $N_{\text{pair}}$ offspring per iteration.

\paragraph{Pair crossover.}
\label{sec:method_crossover}
Crossover produces a hybrid offspring as follows (Fig.~\ref{fig:overview}, step~5):
\begin{enumerate}
    \item \textbf{\emph{Generate.}} A crossover LLM produces the offspring conditioned on the hybridization insight $I_{\times}$ of the same pair (Eq.~\ref{eq:reflection}), which names the two complementary mechanisms and how to blend them:
    \begin{equation}
        h' = \boldsymbol{\pi}_{\text{crossover}}(h_i, h_j, I_{\times}).
        \label{eq:crossover}
    \end{equation}
    \item \textbf{\emph{Build the evaluation cube.}} A cube $\mathcal{B}(\bar{\mathbf{k}}, \rho)$ is centered between the parents, $\bar{\mathbf{k}} = \lfloor (\bar{\boldsymbol{\varphi}}(h_i) + \bar{\boldsymbol{\varphi}}(h_j))/2 \rceil$.
    \item \textbf{\emph{Replace locally.}} The offspring replaces the incumbent specialist in each cell $\mathbf{k}' \in \mathcal{B}$ if $P(h'; X_{\mathbf{k}'}) > P(h_{\mathbf{k}'}; X_{\mathbf{k}'})$.
\end{enumerate}
Since the offspring blends the two parents, its strengths are expected near their regions; local evaluation focuses computation there, and prevents a region-specialized offspring from being dismissed for losing in distant regions.

\paragraph{Region-aware mutation.}
\label{sec:method_mutation}
Mutation specializes a single parent toward a \emph{target neighborhood} of the grid:
\begin{enumerate}
    \item[(i)] \emph{Parent selection.} Heuristics are ranked by grid coverage $n(h) = |\mathcal{C}(h)|$ and the $N_{\text{cand}}$ most-covered are retained; a parent $h$ is sampled under inverse-rank probability $p_r \propto (r+1)^{-1}$, rank $r=1$ being the most-covered, favoring broadly successful heuristics.
    \item[(ii)] \emph{Target cube selection.} $C{=}10$ candidate cubes are built around random filled cells and ranked by mean cell staleness $\sigma(\mathbf{c}_{\mathbf{k}}) = (\nu_{\mathbf{k}} + 1)^{-1}$; the target cube $\mathcal{B}$ is sampled under $p_r \propto (r+1)^{-1}$, rank $r=1$ being the stalest, directing mutation toward least-updated regions.
    \item[(iii)] \emph{Insight retrieval.} From $\mathcal{B}$ we retrieve up to $L{=}2$ unique insights $\hat{\Xi}$, rarest first and then most recent first: rarity surfaces the region's under-used mechanisms rather than the already-dominant one, and recency keeps them aligned with the current specialists rather than superseded ones.
\end{enumerate}
The mutation LLM (Fig.~\ref{fig:overview}, step~6) then generates
\begin{equation}
    h'' = \boldsymbol{\pi}_{\text{mutation}}\big(h, \mathrm{bundle}(\hat{\Xi})\big),
    \label{eq:mutation}
\end{equation}
where $\mathrm{bundle}(\cdot)$ concatenates the retrieved insights into the prompt. As in crossover, the mutant replaces the incumbent specialist in each cell $\mathbf{k} \in \mathcal{B}$ if $P(h''; X_{\mathbf{k}}) > P(h_{\mathbf{k}}; X_{\mathbf{k}})$. Because the parent typically lives outside $\mathcal{B}$, mutation adapts a heuristic strong in one region toward a stale region using that region's insights, enabling cross-regional transfer.

\begin{table*}[t]
\centering
\small
\renewcommand{\arraystretch}{0.98}
\setlength{\extrarowheight}{-0.3pt}
\setlength{\tabcolsep}{14pt}
\resizebox{\textwidth}{!}{%
\begin{tabular}{ll|rGrGr|rGrGrGr}
\bottomrule[0.5mm]
 & & \multicolumn{3}{c|}{\textbf{TSP}} & \multicolumn{4}{c}{\textbf{KP}} \\
\hline
\textbf{Method} & \textbf{Selection} & $n{=}50$ & $n{=}100$ & $n{=}200$ & $n{=}50$ & $n{=}100$ & $n{=}200$ & $n{=}500$ \\
\midrule[0.3mm]
Greedy Construct &  & 21.88 & 25.28 & 27.84 & 0.569 & 0.259 & 0.193 & 0.135 \\
\hline
\multicolumn{9}{c}{LLM-based AHD: \emph{GPT-4o-mini}} \\
\hline
ReEvo            & Top-3  & 13.85 & 17.67 & 20.42 & 0.668 & 0.452 & 0.475 & 0.453 \\
                 & Top-5  & 13.21 & 17.13 & 20.13 & 0.574 & 0.371 & 0.314 & 0.322 \\
                 & Oracle & 12.25 & 16.28 & 19.46 & 0.314 & 0.172 & 0.145 & 0.162 \\
\arrayrulecolor{gray!40}\hline\arrayrulecolor{black}
MCTS-AHD         & Top-3  & 14.86 & 18.50 & 24.62 & 0.493 & 0.234 & 0.202 & 0.252 \\
                 & Top-5  & 11.11 & 15.12 & 21.40 & 0.383 & 0.189 & 0.154 & 0.176 \\
                 & Oracle & 9.04  & 12.86 & 16.27 & 0.313 & 0.156 & 0.126 & 0.118 \\
\arrayrulecolor{gray!40}\hline\arrayrulecolor{black}
PathWise         & Top-3  & 12.93 & 16.28 & 22.16 & 0.434 & 0.201 & 0.188 & 0.212 \\
                 & Top-5  & 10.00 & 13.46 & 18.62 & 0.345 & 0.168 & 0.149 & 0.153 \\
                 & Oracle & 8.41  & 12.22 & 15.62 & 0.288 & 0.142 & 0.120 & 0.105 \\
\arrayrulecolor{gray!40}\hline\arrayrulecolor{black}
EoH-S            & Top-3  & 8.99  & 13.01 & 16.04 & \NAcell & \NAcell & \NAcell & \NAcell \\
                 & Top-5  & 8.26  & 12.17 & 15.28 & \NAcell & \NAcell & \NAcell & \NAcell \\
                 & Oracle & 7.30  & 11.24 & 14.56 & \NAcell & \NAcell & \NAcell & \NAcell \\
\arrayrulecolor{gray!40}\hline\arrayrulecolor{black}
MOSAIC (Ours) & Top-3  & \textbf{8.62} & \textbf{12.05} & \textbf{14.78} & \textbf{0.348} & \textbf{0.184} & \textbf{0.154} & \textbf{0.133} \\
                 & Top-5  & \textbf{7.78} & \textbf{11.27} & \textbf{14.06} & \textbf{0.279} & \textbf{0.157} & \textbf{0.137} & \textbf{0.117} \\
                 & Oracle & \textbf{6.10} & \textbf{9.71} & \textbf{12.72} & \textbf{0.207} & \textbf{0.114} & \textbf{0.098} & \textbf{0.080} \\
\hline
\multicolumn{9}{c}{LLM-based AHD: \emph{GPT-5-nano}} \\
\hline
ReEvo            & Top-3  & 13.23 & 16.78 & 19.93 & 0.554 & 0.309 & 0.347 & 0.414 \\
                 & Top-5  & 12.94 & 16.36 & 19.71 & 0.533 & 0.302 & 0.340 & 0.408 \\
                 & Oracle & 12.04 & 15.38 & 18.72 & 0.504 & 0.284 & 0.323 & 0.383 \\
\arrayrulecolor{gray!40}\hline\arrayrulecolor{black}
MCTS-AHD         & Top-3  & 7.55  & 10.63 & 15.51 & 0.495 & 0.321 & 0.542 & 0.691 \\
                 & Top-5  & 6.23  & 9.07  & 15.34 & 0.333 & 0.212 & 0.254 & 0.309 \\
                 & Oracle & 5.50  & 8.15  & 15.14 & 0.227 & 0.143 & 0.144 & 0.152 \\
\arrayrulecolor{gray!40}\hline\arrayrulecolor{black}
PathWise         & Top-3  & 6.57  & 9.99  & 15.21 & 0.441 & 0.292 & 0.488 & 0.574 \\
                 & Top-5  & 5.61  & 8.62  & 14.27 & 0.303 & 0.195 & 0.234 & 0.266 \\
                 & Oracle & 5.12  & 7.82  & 14.23 & 0.204 & 0.134 & 0.135 & 0.137 \\
\arrayrulecolor{gray!40}\hline\arrayrulecolor{black}
EoH-S            & Top-3  & 6.37  & 11.31 & 15.98 & \NAcell & \NAcell & \NAcell & \NAcell \\
                 & Top-5  & 5.51  & 10.33 & 14.05 & \NAcell & \NAcell & \NAcell & \NAcell \\
                 & Oracle & 4.82  & 8.65  & 12.64 & \NAcell & \NAcell & \NAcell & \NAcell \\
\arrayrulecolor{gray!40}\hline\arrayrulecolor{black}
MOSAIC (Ours) & Top-3  & \textbf{6.01} & \textbf{9.06} & \textbf{12.18} & \textbf{0.183} & \textbf{0.119} & \textbf{0.115} & \textbf{0.177} \\
                 & Top-5  & \textbf{5.03} & \textbf{7.93} & \textbf{10.96} & \textbf{0.145} & \textbf{0.095} & \textbf{0.085} & \textbf{0.118} \\
                 & Oracle & \textbf{3.15} & \textbf{5.74} & \textbf{8.81} & \textbf{0.062} & \textbf{0.042} & \textbf{0.042} & \textbf{0.051} \\
\toprule[0.5mm]
\end{tabular}%
}
\caption{Heuristic generation on TSP and KP: mean optimality gap (\%) over runs on 200 (TSP) and 500 (KP) test instances per size. LKH-3~\citep{lkh3} and OR-Tools~\citep{ortools} provide the optimal values for TSP and KP. Best value per column and selection level within each LLM-backbone block in bold; \NAcell{} marks a method not available for the problem.}
\label{tab:main_results}
\end{table*}

\subsection{Portfolio Extraction at Test Time}
\label{sec:method_select}

The final archive stores many distinct specialists $\mathcal{H}_{\mathcal{G}}$. We extract a heuristic portfolio $A_k \subseteq \mathcal{H}_{\mathcal{G}}$ once, offline, by greedy complementary selection on held-out validation instances $\mathcal{D}_{\text{val}}$: starting from $A_0 = \varnothing$, we repeatedly add the heuristic that most reduces the mean best-of-set cost,
\begin{equation}
    A_{j} = A_{j-1} \cup \Big\{ \operatorname*{arg\,min}_{h \in \mathcal{H}_{\mathcal{G}} \setminus A_{j-1}} \frac{1}{|\mathcal{D}_{\text{val}}|} \sum_{x \in \mathcal{D}_{\text{val}}} \min_{h' \in A_{j-1} \cup \{h\}} f(h'(x)) \Big\}.
    \label{eq:portfolio}
\end{equation}
The Top-$k$ portfolio is thus the length-$k$ prefix of a single ranking. Complementarity is enforced by the selection criterion, which credits a heuristic only where it improves the current set, and is supplied by the co-evolution itself: instances are evolved to separate heuristics, and a specialist survives in the archive only by winning feature-space regions.

\section{Experiments}
\label{sec:experiments}

We evaluate MOSAIC on two tasks: \textbf{heuristic generation}, where the heuristics of the co-evolved archive are compared against LLM-based AHD baselines on the Traveling Salesman Problem (TSP), the Knapsack Problem (KP), and the Capacitated Vehicle Routing Problem (CVRP); and \textbf{instance generation}, where the discriminative instances evolved by our framework are compared against evolutionary instance-generation baselines on TSP. Experimental details, metric definitions, and the CVRP results are provided in Appendices~\ref{app:experimental_details}, \ref{app:metrics}, and~\ref{app:cvrp}, respectively.

\paragraph{Benchmarks.}
Heuristics are constructive on all three problems, building a solution sequentially. Test sets span structurally distinct families: uniform, grid-like, ring, and mixture city layouts for TSP; uncorrelated, weakly, strongly, and inverse-correlated, and subset-sum value--weight relations for KP (Appendix~\ref{app:benchmarks}). TSP has 200 test instances per size $n \in \{50, 100, 200\}$, KP 500 per size $n \in \{50, 100, 200, 500\}$. Methods train at a single size ($n{=}50$ for TSP, $n{=}100$ for KP); the remaining sizes and structures measure generalization across \emph{problem scale} and \emph{instance structure}.

\paragraph{Baselines.}
For heuristic generation, we compare against the LLM-based AHD methods \textbf{ReEvo}~\citep{ye2024reevo}, \textbf{MCTS-AHD}~\citep{zheng2025montecarlotreesearch}, \textbf{PathWise}~\citep{gungordu2026pathwise}, and \textbf{EoH-S}~\citep{eohs}, and the manually designed \textbf{Greedy Construct} heuristic, with GPT-4o-mini and GPT-5-nano as LLM backbones. Every LLM-based method is run 9 times (3 seeds, 3 runs per seed) under the same budget of 300 generated heuristics.

For instance generation, we compare against the evolutionary algorithm (EA) baselines \textbf{\boldmath$(\mu{,}\lambda)$-EA}~\citep{smith2011discovering}, \textbf{\boldmath$(\mu{+}1)$-CM-EA}~\citep{bossek2019foga}, and \textbf{QD-EA}~\citep{bossek2022gecco}. We evaluate three variants of our generator that isolate the contributions of the LLM and the evolutionary loop: \textbf{Ours (Classical EA)} replaces the LLM-designed instance-generation function $g_{\text{inst}}$ with the $(\mu{,}\lambda)$-EA operators, \textbf{Ours (LLM-Direct)} generates the instance set in a single LLM call with no evolutionary loop, and \textbf{Ours (LLM-Guided EA)} is the framework's default generator (Section~\ref{sec:method_isa}). All methods run under the same heuristic-evaluation budget, with GPT-5-nano.

\paragraph{Evaluation protocol.}
For heuristic generation, each run produces a heuristic pool. \emph{Top-$k$} reports the per-instance best objective among the run's $k$ top-ranked heuristics; \emph{Oracle} reports the per-instance best over the run's full heuristic pool, an upper bound on any selection from that pool. We report the mean optimality gap over runs.

For instance generation, each method's evolved instance set is scored on two axes: \emph{feature-space coverage}, via the coverage ratio (CR), the fraction of the feature grid it fills; and \emph{algorithmic performance} against five traditional constructive heuristics, via the mean optimality gap (Gap) and the instance discrimination score (Discrim), the gap difference between the best and worst heuristic, averaged over instances.

\subsection{Heuristic Generation Results}
\label{sec:exp_heuristic}

\begin{figure}[t]
    \centering
    \includegraphics[width=\columnwidth]{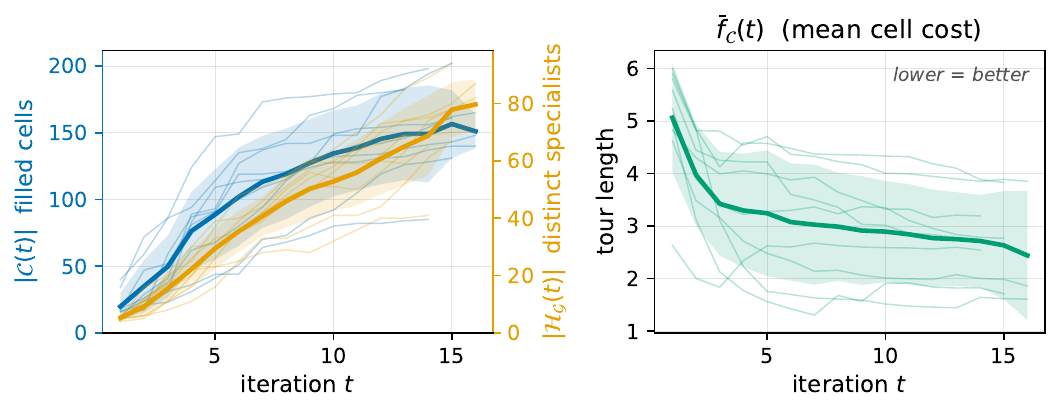}
    \caption{Archive dynamics on TSP over the co-evolutionary iterations $t$, aggregated across 9 independent runs. \textbf{Left:} filled grid cells $|\mathcal{C}(t)|$ (blue, left axis) and distinct specialists $|\mathcal{H}_{\mathcal{G}}(t)|$ (orange, right axis). \textbf{Right:} mean per-cell cost $\bar{f}_{\mathcal{C}}(t) = \frac{1}{|\mathcal{C}|}\sum_{\mathbf{k} \in \mathcal{C}} f(h_{\mathbf{k}}; X_{\mathbf{k}})$ (green), the negative of the mean per-cell performance maximized by Eq.~\ref{eq:heuristic_obj}.}
    \label{fig:archive_dynamics}
\end{figure}

MOSAIC attains the lowest gap in Table~\ref{tab:main_results} at every selection level, size, and LLM backbone. The Oracle rows show that the archive is a stronger heuristic pool than any baseline's, with gaps up to $35\%$ lower on TSP and $70\%$ lower on KP than the best baseline's. At Top-3 and Top-5, the gap is up to $22\%$ lower on TSP and $67\%$ on KP, and with GPT-5-nano the Top-5 gap is lower than every baseline's \emph{Oracle} gap on KP at all sizes and on TSP at $n{=}200$. Gains persist out of distribution and remain substantial at the largest test sizes, including $n{=}200$ on TSP and $n{=}500$ on KP. The gains extend to CVRP, where the gap is up to $35\%$ lower than the best baseline's (Appendix~\ref{app:cvrp}, Table~\ref{tab:cvrp_main}). Figure~\ref{fig:archive_dynamics} traces the source of this heuristic-pool quality: coverage $|\mathcal{C}(t)|$ and the number of distinct specialists grow throughout the run as the co-evolved instances keep expanding the filled set, while the mean per-cell cost falls as specialists improve, with upticks when newly discovered, harder regions enter the average. Table~\ref{tab:instgen_main} independently confirms the hardness of these co-evolved instances. Rising coverage and falling cost are the two axes of quality--diversity: the archive simultaneously diversifies and improves. Final-archive visualizations for TSP and KP show the grid partitioned among many region specialists (Appendix~\ref{app:visualizations}, Figure~\ref{fig:winner_map}).

\begin{table}[t]
\centering
\footnotesize
\renewcommand{\arraystretch}{0.95}
\setlength{\extrarowheight}{-0.3pt}
\setlength{\tabcolsep}{0.5pt}
\resizebox{\columnwidth}{!}{%
\begin{tabular}{l ccc}
\toprule
 & \multicolumn{1}{c}{\scriptsize\textbf{Feature-Space Coverage}} & \multicolumn{2}{c}{\scriptsize\textbf{Algorithmic Performance}} \\
\cmidrule(lr){2-2} \cmidrule(l){3-4}
\textbf{Method} & \textbf{CR(\%)}$\,\uparrow$ & \textbf{Gap(\%)}$\,\uparrow$ & \textbf{Discrim(\%)}$\,\uparrow$ \\
\midrule
$(\mu{,}\lambda)$-EA    & $3.43{\pm}0.41$  & $146.2{\pm}1.1$ & $537{\pm}9$   \\
$(\mu{+}1)$-CM-EA & $3.49{\pm}0.10$  & $147.7{\pm}1.9$ & $543{\pm}12$  \\
QD-EA                   & $25.12{\pm}1.95$ & $147.6{\pm}2.7$ & $554{\pm}14$  \\
\arrayrulecolor{gray!40}\cmidrule{1-4}\arrayrulecolor{black}
Ours (Classical EA)     & $12.42{\pm}0.30$ & $150.2{\pm}1.5$ & $551{\pm}9$   \\
Ours (LLM-Direct)       & $28.20{\pm}1.85$ & $140.6{\pm}8.5$ & $525{\pm}40$  \\
\arrayrulecolor{gray!40}\cmidrule{1-4}\arrayrulecolor{black}
Ours (LLM-Guided EA)    & $\mathbf{33.45{\pm}0.87}$ & $\mathbf{182.1{\pm}6.5}$ & $\mathbf{744{\pm}36}$ \\
\bottomrule
\end{tabular}%
}
\caption{Instance generation on TSP with $n{=}50$ cities at a budget of $5{\times}10^3$ heuristic evaluations (mean $\pm$ std over 3 seeds); best value per column in bold; last row is the default.}
\label{tab:instgen_main}
\end{table}

\subsection{Instance Generation Results}
\label{sec:exp_instance}

We compare the instance generators at a budget of $5{\times}10^3$ heuristic evaluations. Table~\ref{tab:instgen_main} shows our generator is the only one winning both axes: it attains the highest feature-space coverage \emph{and} the hardest, most discriminating instances, while baselines trade one for the other. It covers $33\%$ more of the feature grid than QD-EA while producing instances $23\%$ harder and $34\%$ more discriminating; full coverage is unattainable, as some feature combinations admit no instance. The advantage persists at $n{=}100$, where the hardness and discrimination margins grow to $48\%$ and $60\%$; the ranking already holds at one fifth of the budget (Appendix~\ref{app:instgen_extended}), and Figures~\ref{fig:pair_instances} and~\ref{fig:pair_scatter} (Appendix~\ref{app:visualizations}) illustrate generated instances and the feature-space regions they define.

\begin{table}[t]
\centering
\footnotesize
\renewcommand{\arraystretch}{0.90}
\setlength{\tabcolsep}{5pt}
\begin{tabular}{l r}
\toprule
\textbf{Ablation} & \textbf{AVG (\%)}$\,\downarrow$ \\
\midrule
MOSAIC (default)                  & \textbf{9.57} \\
\arrayrulecolor{gray!40}\midrule\arrayrulecolor{black}
\multicolumn{2}{l}{\emph{QD grid archive}} \\
\quad W/o QD grid (flat population)     & 13.12 \\
\quad Global offspring evaluation       & 10.73 \\
\quad Random parent pairs               & 10.51 \\
\quad Random mutation targeting         & 12.45 \\
\quad Static instance set               & 10.24 \\
\quad W/o decision-tree filter          & 10.35 \\
\arrayrulecolor{gray!40}\midrule\arrayrulecolor{black}
\multicolumn{2}{l}{\emph{Operator mix}} \\
\quad W/o crossover                     & 10.01 \\
\quad W/o mutation                      & 15.28 \\
\arrayrulecolor{gray!40}\midrule\arrayrulecolor{black}
\multicolumn{2}{l}{\emph{Contrastive multi-directional feedback}} \\
\quad Uni-directional feedback          & 10.46 \\
\quad Crossover w/o insight             & 12.29 \\
\quad Mutation w/o insights             & 11.91 \\
\arrayrulecolor{gray!40}\midrule\arrayrulecolor{black}
\multicolumn{2}{l}{\emph{Insight retrieval}} \\
\quad Most-widespread insight retrieval & 10.08 \\
\quad Random insight retrieval          & 10.68 \\
\bottomrule
\end{tabular}
\caption{Ablations of MOSAIC on TSP. AVG is the mean optimality gap (\%) over Top-3, Top-5, and Oracle at $n{=}50$ and $n{=}100$, over 6 runs per ablation on the first 100 test instances of each size; lower is better.}
\label{tab:ablation_main}
\end{table}

\subsection{Ablation Study}
\label{sec:exp_ablation}

We conduct an ablation study on TSP with the GPT-4o-mini LLM backbone to evaluate the core components of MOSAIC. Table~\ref{tab:ablation_main} reports AVG, the mean optimality gap over Top-3, Top-5, and Oracle at test sizes $n{\in}\{50,100\}$. The default configuration attains the lowest AVG. Additional ablations, covering parameter sweeps and the results per selection level and size, are provided in Appendix~\ref{app:extended_ablations}.

\paragraph{QD grid archive.}
To separate the archive from prompts and operators, the w/o-QD-grid ablation replaces it with a flat population of 20 kept by global elitism, removing region specialists, cube-based local evaluation, and spatially-grounded insight storage. AVG worsens by 37\%, since a heuristic winning one feature-space region but losing on average is discarded, leaving a heuristic pool of near-duplicates. The remaining ablations remove the grid's guidance to the operators. Evaluating offspring globally instead of on local cubes worsens AVG by 12\%, random parent pairs by 10\%, and random mutation targeting by 30\%, so feature-distant pairing and staleness-directed mutation are the grid signals operators depend on most. Keeping the stored instances fixed at $\mathcal{D}_{\text{init}}$ worsens AVG by 7\%, since the evolved discriminative instances never enter the archive and specialists adapt only to the initial distribution. Removing the decision-tree filter worsens AVG by 8\%, since unvalidated instances and insights accumulate conflicting winners within cells.

\paragraph{Operator mix.}
Removing mutation worsens AVG by 60\%, the largest degradation. The heuristic-generation budget flows only through crossover, which recombines existing heuristics but neither uses region-aware specialization insights nor introduces new mechanisms, so all runs collapse to a single heuristic and terminate early, without developing region specialists. Removing crossover worsens AVG by only 5\%. Mutation sustains the search; crossover contributes hybrid offspring recombining distant specialists' mechanisms.

\paragraph{Contrastive multi-directional feedback.}
Replacing the three contrastive insights with a uni-directional better/worse hint worsens AVG by 9\%, so the insights' directional content carries signal beyond the winner's identity. Removing the insights from the operator prompts has a larger effect: mutation w/o insights worsens AVG by 24\% and crossover w/o insight by 28\%, exceeding the 5\% degradation of removing crossover entirely. Offspring generated without insights still combine or modify their parents without regional guidance, enter grid cells through cell-level replacement, and are later selected as parents, injecting noise into the grid that propagates through pair analysis, reflection, and the next offspring.

\paragraph{Insight retrieval.}
Retrieving insights by rarity, then recency prevents a feedback loop where the most widespread insight is retrieved and regenerates the mechanism that produced it. Retrieving the most widespread insight first worsens AVG by 5\%, and random retrieval worsens it by 12\%, so the retrieval policy contributes beyond insight availability.

\paragraph{Instance generation.}
Table~\ref{tab:instgen_main} also serves as the instance-generator ablation: Ours (Classical EA) removes the LLM operator, Ours (LLM-Direct) removes the evolutionary loop, and only their combination attains the highest feature-space coverage and the hardest, most discriminating instances.

\section{Conclusion}
\label{sec:conclusion}

We presented MOSAIC, a grid-based co-evolutionary framework that couples discriminative instance generation with specialist heuristic generation inside a QD archive indexed by instance features. Contrastive multi-directional reflection anchors insights to each heuristic's winning regions, and the archive accumulates specialists, instances, and insights across iterations. Experiments across COPs showed consistent gains over LLM-based AHD methods, while the co-evolved instances surpassed evolutionary instance-generation baselines in coverage and discrimination.

\bibliography{bib/algo_selection,bib/algorithm_surrogate_models,bib/intro_prelim_relatedworks,bib/isa,bib/llm_heuristic,bib/llm_heuristic_instance,bib/llm_reasoning,bib/or_heuristic}

\clearpage

\appendix

\section{Extended Related Work}
\label{app:related}

\subsection{AHD and Hyper-Heuristics}
Because most COPs in logistics, scheduling, and design are NP-hard, heuristics are the practical route to high-quality solutions at scale~\citep{Ausiello2012}. Metaheuristics such as tabu search~\citep{Glover1990}, simulated annealing~\citep{Kirkpatrick1983}, and iterated local search~\citep{Lourenco2003} deliver strong solutions under limited budgets, yet they rely on hand-designed, solver-specific components, keeping heuristic development expert-driven, costly, and poorly transferable across problem variants~\citep{Desale2015,Choong2018,pillay2018}.

Hyper-heuristics search over heuristics rather than solutions~\citep{cowling2000,burke2019hyper,dokeroglu2024}: selection variants choose among predefined low-level heuristics, while generation variants construct new heuristics, most prominently via genetic programming (GP) over syntax trees~\citep{Langdon2013,fukunaga2002,Sabar2013,branke2016,Kieffer2020}; reinforcement learning further treats the choice among heuristics as a sequential decision task~\citep{Choong2018,zhang2022}. GP-based generation, however, is confined to human-defined primitives, restricting the expressiveness of the discovered heuristics~\citep{gpahd}. MOSAIC is a generation approach without this constraint---heuristics are free-form LLM-written programs---and its QD archive maintains many region specialists instead of one heuristic per problem class.

\subsection{LLM-based AHD}
LLMs combine code generation with reasoning~\citep{chen2021evaluatinglargelanguagemodels,yang-etal-2024-llms}, and structured approaches strengthen the latter through planning with world models~\citep{hao-etal-2023-reasoning,xiong2025swap}, structured multi-level modeling~\citep{xiong2026enhancing}, adaptive control of search-augmented reasoning~\citep{xiong2026scaling}, and translation to formal representations~\citep{yang2024harnessing}. Within evolutionary computation, LLMs act as variation operators that produce offspring conditioned on selected parents~\citep{lehman2024,meyerson,lange2024,wu2024}, a paradigm applied to program synthesis~\citep{evoprompting,hemberg2024,Bradley2024} and prompt evolution~\citep{fernando2024,guo2024connecting,griebhaber-etal-2025-toolbox}. LLM-as-optimizer methods instead refine solutions of individual instances in context~\citep{yang2024large,guo2024towards,liullmoptimizer}; they do not produce reusable algorithms and struggle on large search spaces~\citep{llmasevoptimizers2024,zhao2023large}.

LLM-based AHD integrates these capabilities into iterative search over executable heuristics~\citep{liu2023algorithm}. FunSearch~\citep{funsearch} couples an island-based evolutionary loop with LLM code mutation; EoH~\citep{eoh} co-evolves natural-language ``thoughts'' with their implementations; ReEvo~\citep{ye2024reevo} adds short- and long-term verbal reflections~\citep{shinn2023reflexion,madaan2023selfrefine}; HSEvo~\citep{hsevo} manages diversity via harmony search~\citep{Shi2013}; MCTS-AHD~\citep{zheng2025montecarlotreesearch} and Planning-of-Heuristics~\citep{wang2025planningheuristicsstrategicplanning} organize candidates in Monte Carlo search trees~\citep{swiechowski2023mcts}; and PathWise~\citep{gungordu2026pathwise} plans heuristic evolution over an entailment graph that records derivation history, coordinating policy, world-model, and critic agents for state-aware search. CALM~\citep{huang2025calmcoevolutionalgorithmslanguage} fine-tunes the LLM backbone with reinforcement learning during the search, and HeurAgenix~\citep{yang2025heuragenixleveragingllmssolving} pairs heuristic evolution with LLM-guided per-state heuristic selection at solve time. Closest to our motivation, EoH-S~\citep{eohs} evolves a small complementary heuristic set rather than a single heuristic, yet candidates are still evaluated on a fixed training set, so the set specializes only to the regimes that dataset contains. Analyses of LLM-driven search further report weak, noisy fitness signals and redundant exploration~\citep{vsteincodeevolutiongraphs,liu2025fitnesslandscapelargelanguage,nie2024importancedirectionalfeedbackllmbased}, and systematic benchmarking finds that LLM-evolved heuristics generalize poorly beyond the distribution they were evolved on~\citep{sim2025hypebenchmarkingllmevolvedheuristics}.

EALG~\citep{duan2025ealgevolutionaryadversarialgeneration} co-evolves instances and heuristics adversarially, but targets hardness amplification alone: instances are regenerated each round, never stored or analyzed, and the resulting heuristics are general-purpose rather than region-specialized. InstSpecHH~\citep{zhang2025llmdriveninstancespecificheuristicgeneration} partitions a benchmark into feature-defined subclasses and evolves one heuristic per subclass, but the instances are pre-generated and each subclass is searched independently, without transferring knowledge between neighboring regions. Neither couples instance generation and heuristic generation under an explicit instance-space view that simultaneously rewards feature-space coverage and algorithmic discrimination. In contrast, MOSAIC stores decision-tree-validated instances, specialists, and localized insights in one persistent QD archive, transfers mechanisms across regions through insight-driven crossover and mutation, and extracts a complementary portfolio from the final archive.

\subsection{Quality--Diversity and Instance Space Analysis}
QD algorithms replace the search for a single optimum with an archive of diverse, high-performing solutions~\citep{pugh2016qdnature}. MAP-Elites~\citep{mouret2015mapelites} discretizes a behavior space into a grid and keeps the best solution per cell; multi-task variants treat each cell as a related task whose elites seed one another~\citep{mouret2020multitaskqd}, and LLMatic combines QD archives with LLM-generated candidates for neural architecture search~\citep{nasir2024llmatic}. MOSAIC adopts this machinery but repurposes the axes: cells are indexed by structural \emph{instance} features rather than solution behavior descriptors, so the grid partitions the instance space, and each cell stores a specialist together with the representative instances and localized insights that justify it---the cells-as-tasks view of multi-task QD instantiated over an instance space.

ISA characterizes algorithms by where in a structural feature space they succeed, delineating footprints of strength and weakness~\citep{SMITHMILES2012875,SMITHMILES201412instanceinisaspace}, with applications to knapsack~\citep{SMITHMILES2021105184} and vehicle routing~\citep{gouvea2025instancespaceanalysiscapacitated} problems; TSP feature sets with analytically known bounds support automated algorithm selection~\citep{HEINS2023123,10.1145/3450218.3477308}. The perspective is rooted in algorithm selection~\citep{aslib,pulatov2022opening,asllm} and algorithm portfolios, which compose complementary solvers so that each instance is served by its best member~\citep{gomes2001algorithm,xu2010hydra,tang2021few}; the NFL theorem provides the theoretical backdrop~\citep{wolpert1997nfl}, and empirical studies confirm that heuristic dominance is regime-dependent~\citep{datasetconfig1,bppfeatures}. All of these treat the algorithm set as frozen---analysis and selection happen after design---whereas MOSAIC uses ISA-style features during design, so specialists and their winning regions emerge together.

A complementary line evolves the instances themselves: to expose what makes TSP hard for particular solvers~\citep{miles2010tspinstanceevol,smith2011discovering}, to fill underrepresented regions of the feature space~\citep{SMITHMILES2015evolveinstance}, to diversify instance structure through creative mutation operators~\citep{bossek2019foga} or MAP-Elites~\citep{bossek2022gecco}, and to generate diverse, discriminatory knapsack instances via novelty search~\citep{10.1007/978-3-031-14714-2_16}. These generators evolve instances against a fixed solver pair, offline, for analysis or benchmarking; our instance-generation experiments (Section~\ref{sec:exp_instance}) compare against this family under a shared budget. In MOSAIC, instances are evolved to discriminate between continually changing heuristic pairs, validated by a decision tree, stored in the archive as evaluation data, and addressed by newly specialized heuristics---closing the loop between instance generation and heuristic generation.

\section{Details of Benchmark Problems}
\label{app:benchmarks}

This section gives the definition and mathematical formulation of each problem, together with the structural instance families used to build the test sets $\mathcal{D}_{\text{test}}$. All three problems are solved constructively: a heuristic builds a solution sequentially, one decision at a time. Dataset sizes and counts are given in Appendix~\ref{app:experimental_details}.

\subsection{Traveling Salesman Problem (TSP)}

\textbf{Definition.}
The TSP seeks the shortest Hamiltonian tour, visiting every city exactly once and returning to the starting city.

\textbf{Formulation.}
On a complete graph $G=(V,E)$ with $V=\{1,\dots,n\}$ and edge costs $c_{ij} \geq 0$, let $x_{ij}\in\{0,1\}$ indicate whether the tour traverses edge $(i,j)$:
\begin{equation}
\begin{aligned}
&\text{minimize}\quad \textstyle\sum_{i\in V}\sum_{j\in V} c_{ij}x_{ij}, \\
&\text{s.t.}\quad \textstyle\sum_{j\in V}x_{ij}=1,\;\; \textstyle\sum_{i\in V}x_{ij}=1, \quad \forall i,j \in V,\\
&\phantom{\text{s.t.}\quad} \textstyle\sum_{i\in S}\sum_{j\in S}x_{ij}\le |S|-1, \quad \forall S\subset V,\; 2\le |S| \le n-1.
\end{aligned}
\end{equation}
In our instances, cities are points in the unit square $[0,1]^2$ and $c_{ij}$ is the Euclidean distance. A constructive heuristic selects the next city given the current city, the destination, the unvisited set, and the distance matrix.

\textbf{Test instance families.}
Each test set contains four structural families in equal proportion, each generated by a family-specific sampler in the unit square with per-instance randomized structural parameters, so instances within a family differ structurally rather than being near-copies:
\begin{itemize}
    \item \emph{Uniform}: all $n$ cities drawn i.i.d.\ from $U[0,1]^2$.
    \item \emph{Grid}: cities placed on $n$ nodes sampled without replacement from a regular $\lceil\sqrt{n}\rceil \times \lceil\sqrt{n}\rceil$ lattice, then perturbed with Gaussian jitter whose standard deviation is drawn from $U[0,\, 0.5/\lceil\sqrt{n}\rceil]$.
    \item \emph{Ring}: cities placed on 1--3 concentric circles centered at $(0.5, 0.5)$ with radii drawn from $U[0.12, 0.48]$; each city takes a uniform angle on a random circle with radial Gaussian noise of standard deviation $0.02$.
    \item \emph{Mixture}: $\lfloor n/2 \rfloor$ uniform cities combined with cities from a clustered layout of 2--8 Gaussian blobs with uniform centers and a shared standard deviation drawn from $U[0.03, 0.10]$, producing heterogeneous density.
\end{itemize}
All coordinates are clipped to $[0,1]^2$, and optimal values are computed by LKH-3~\citep{lkh3}.

\subsection{Knapsack Problem (KP)}

\textbf{Definition.}
The KP selects the item subset of maximum total value under a single weight-capacity constraint.

\textbf{Formulation.}
With item values $v_i \geq 0$, weights $w_i \geq 0$, and capacity $W$, let $x_i\in\{0,1\}$ indicate whether item $i$ is selected:
\begin{equation}
\begin{aligned}
&\text{maximize}\quad \textstyle\sum_{i=1}^{n} v_i x_i, \\
&\text{s.t.}\quad \textstyle\sum_{i=1}^{n} w_i x_i \le W,\quad x_i\in\{0,1\}.
\end{aligned}
\end{equation}
In our instances, weights and values lie in $[0,1]$ and the capacity is fixed by the instance size, $W{=}12.5$ for $n{=}50$ and $W{=}25$ otherwise, following ReEvo~\citep{ye2024reevo}. A constructive heuristic selects the next item given the remaining capacity and the weights and values of the remaining items.

\textbf{Test instance families.}
Each test set contains five classic value--weight correlation regimes in equal proportion, rescaled to $[0,1]$:
\begin{itemize}
    \item \emph{Uncorrelated}: $w_i, v_i \sim U(0,1)$ drawn independently.
    \item \emph{Weakly correlated}: $v_i = w_i + U(-0.25, 0.25)$, so value tracks weight with noise.
    \item \emph{Strongly correlated}: $v_i = w_i + 0.2 + U(-0.01, 0.01)$, compressing value-to-weight ratios so that greedy ratio ordering is weakly informative.
    \item \emph{Inverse-correlated}: $v_i = 1 - w_i + U(-0.1, 0.1)$, making light items the valuable ones.
    \item \emph{Subset-sum}: $v_i = w_i + U(-0.02, 0.02)$, so all ratios are close to one and the problem reduces to filling the capacity exactly.
\end{itemize}
All values are clipped to $[0,1]$. Every instance additionally draws a weight scale $s \sim U(0.6, 1.4)$ applied to its weights (clipped to $[0,1]$), so capacity tightness $W/\sum_i w_i$ varies across instances within each family. Optimal values are computed with OR-Tools~\citep{ortools}.

\subsection{Capacitated Vehicle Routing Problem (CVRP)}

\textbf{Definition.}
The CVRP seeks a set of vehicle routes of minimum total length, each starting and ending at a common depot, such that every customer is visited exactly once and the total demand served on each route does not exceed the vehicle capacity.

\textbf{Formulation.}
On a complete graph $G=(V,E)$ with $V=\{0,1,\dots,n\}$, where $0$ is the depot and each customer $i$ carries a demand $q_i > 0$, vehicles are identical with capacity $Q$, edge costs are $c_{ij} \geq 0$, and $x_{ij}\in\{0,1\}$ indicates whether edge $(i,j)$ is traversed:
\begin{equation}
\begin{aligned}
&\text{minimize}\quad \textstyle\sum_{(i,j)\in E} c_{ij}x_{ij}, \\
&\text{s.t.}\quad \textstyle\sum_{j\in V}x_{ij}=2, \quad \forall i \in V\setminus\{0\},\\
&\phantom{\text{s.t.}\quad} \textstyle\sum_{i\in S}\sum_{j\in V\setminus S}x_{ij}\ \geq\ 2\,\big\lceil \textstyle\sum_{i\in S} q_i / Q \big\rceil,
\end{aligned}
\end{equation}
where the second constraint, imposed for every nonempty customer subset $S \subseteq V\setminus\{0\}$, jointly forbids routes disconnected from the depot and enforces the capacity limit. In our instances, the depot and the customers are points in the unit square $[0,1]^2$ and $c_{ij}$ is the Euclidean distance; demands are integers and every instance draws its own vehicle capacity, clamped to at least the maximum demand plus one, so capacity tightness varies across instances. A constructive heuristic selects the next customer given the current position, the depot, the feasible unvisited customers, the remaining vehicle capacity, the demands, and the distance matrix; when no unvisited customer fits the remaining capacity, the route closes and a new vehicle starts from the depot.

\textbf{Test instance families.}
Each test set contains four structural families in equal proportion, each generated by a family-specific sampler with per-instance randomized parameters and the depot drawn uniformly in the unit square:
\begin{itemize}
    \item \emph{Uniform}: customers drawn i.i.d.\ from $U[0,1]^2$ with demands $q_i \sim U\{1,\dots,10\}$ and capacity $Q \sim U\{25,\dots,150\}$.
    \item \emph{Clustered}: customers in 2--6 Gaussian blobs with uniform centers and a shared standard deviation drawn from $U(0.03, 0.10)$; demands and capacity as in the uniform family.
    \item \emph{Depot-correlated}: uniform customers whose demands grow with distance from the depot, assigned by distance rank with Gaussian noise and clipped to $\{1,\dots,10\}$; $Q \sim U\{30,\dots,80\}$.
    \item \emph{Inverse depot-correlated}: as the previous family with the ranking inverted, so customers near the depot carry the heaviest demands; $Q \sim U\{30,\dots,80\}$.
\end{itemize}
All coordinates lie in the unit square, and optimal values are computed by LKH-3~\citep{lkh3}.

\section{Experimental Details}
\label{app:experimental_details}

\subsection{Dataset Configuration}
Table~\ref{tab:dataset_config} summarizes the datasets. The initial dataset $\mathcal{D}_{\text{init}}$ consists of uniformly sampled instances at the training size; it seeds our grid archive and is the same training instance set used by the LLM-based AHD baselines. Because the baselines do not generate instances and can only adapt to this fixed set, we run every method under three training-dataset seeds (3 runs per seed): each method adapts to three different instance draws, so no comparison hinges on a single training set. The validation instances $\mathcal{D}_{\text{val}}$ used for portfolio extraction are uniformly sampled at the training size. The test sets $\mathcal{D}_{\text{test}}$ span the structural families in equal proportion per size.

\begin{table}[t]
\centering
\small
\setlength{\tabcolsep}{5pt}
\begin{tabular}{l cc cc}
\toprule
 & \multicolumn{2}{c}{$\mathcal{D}_{\text{init}}$} & \multicolumn{2}{c}{$\mathcal{D}_{\text{test}}$} \\
\cmidrule(lr){2-3} \cmidrule(l){4-5}
\textbf{Problem} & Size & \#Inst. & Size & \#Inst. \\
\midrule
TSP  & 50  & 64 & \{50, 100, 200\}      & 200 \\
KP   & 100 & 64 & \{50, 100, 200, 500\} & 500 \\
CVRP & 50  & 64 & \{50, 100\}           & 200 \\
\bottomrule
\end{tabular}
\caption{Dataset configuration. $\mathcal{D}_{\text{init}}$ is also the training instance set of the LLM-based AHD baselines; the validation instances $\mathcal{D}_{\text{val}}$ share its instance size. $\mathcal{D}_{\text{test}}$ counts are per size.}
\label{tab:dataset_config}
\end{table}

\subsection{MOSAIC Configuration}
\label{app:ours_config}
All framework hyperparameters and their values are listed in Table~\ref{tab:hyperparams}; the same configuration is used for all three problems and both LLM backbones. The grid features are drawn from standard instance feature groups, distance-distribution statistics and nearest-neighbor-graph statistics for TSP~\citep{HEINS2023123}, value--weight statistics for KP, and capacity--demand statistics for CVRP:
\begin{itemize}
    \item TSP uses \texttt{fraction\_of\_distinct\_distances} $\times$ \texttt{n\_strong}: the fraction of pairwise city distances that are distinct (a distance-distribution feature), and the number of strongly connected components of the directed 3-nearest-neighbor graph (a nearest-neighbor-graph feature).
    \item KP uses \texttt{vw\_correlation} $\times$ \texttt{capacity\_tightness}: the Pearson correlation between item values and weights, and the fraction of the total item weight that fits into the capacity.
    \item CVRP uses \texttt{capacity\_tightness} $\times$ \texttt{demand\_depot\_corr}: the share of the total customer demand that a single vehicle load covers, $\min(Q / \sum_i q_i, 1)$, and the Pearson correlation between customer demands and their distances to the depot.
\end{itemize}

Each axis uses the analytic bounds $[\ell_i, u_i]$ of its feature (Section~\ref{sec:prelim_features}). For TSP at its training size $n{=}50$, \texttt{fraction\_of\_distinct\_distances} is continuous on $[4/(n(n-1)),\,1]$ with $R{=}10$ bins and \texttt{n\_strong} is integer-valued with range $[1, n-3]$~\citep{HEINS2023123}, one cell per value, giving $|\mathcal{G}| = 10 \times 47 = 470$ cells. All KP and CVRP grid features are continuous and size-invariant, \texttt{vw\_correlation} and \texttt{demand\_depot\_corr} on $[-1, 1]$ and \texttt{capacity\_tightness} on $[0, 1]$, giving $|\mathcal{G}| = 10 \times 10 = 100$ cells for each problem.

\begin{table}[h]
\centering
\small
\begin{tabular}{@{}lll@{}}
\toprule
\textbf{Symbol} & \textbf{Description} & \textbf{Default} \\
\midrule
$E_{\max}$ & Heuristic-generation budget & 300 \\
$N_{\text{pair}}$ & Pairs analyzed per iteration & 10 \\
$N_{\text{init}}$ & Initial heuristic candidates & 20 \\
$d$ & Number of features (grid dims) & 2 \\
$R$ & Bin count for continuous features & 10 \\
$M$ & Max instances per cell & 3 \\
$\rho$ & Cube fraction of each axis & 0.4 \\
$T$ & Min filled cells per cube & 64 \\
$N_{\text{cand}}$ & Candidates kept for inverse-rank sampling & 20 \\
$L$ & Max insights per mutation prompt & 2 \\
$C$ & Candidate cubes for staleness & 10 \\
$N_{\text{inst}}$ & Instances per evolution & 50 \\
$G$ & Generations for instance EA & 10 \\
$S$ & Stagnation limit & 5 \\
$D_{\max}$ & Max decision-tree depth & 3 \\
$\alpha$ & Min leaf size as fraction of $|X_{\text{all}}|$ & 0.05 \\
\bottomrule
\end{tabular}
\caption{Hyperparameters of the framework.}
\label{tab:hyperparams}
\end{table}

\subsection{Baseline Configuration}
\paragraph{Heuristic generation.}
All LLM-based baselines run with their original algorithmic configurations under the same generated-heuristic budget ($300$) and the same 60-second per-heuristic time limit as MOSAIC, with 9 runs each (3 dataset seeds, 3 runs per seed); occasional failed runs are excluded, leaving at least 6 evaluated runs per method. Top-$k$ and Oracle operate on the heuristic pool each method itself maintains, in that method's own ranking. For ReEvo the heuristic pool is the population at the end of the run, ranked by training objective. For MCTS-AHD it is the heuristics of the search tree, ranked by node $Q$ value. For PathWise it is the heuristics of the final entailment graph, ranked by training objective. For EoH-S it is the surviving population set in its set-survival order: the heuristic with the best mean training score first, each subsequent one chosen by its marginal improvement to the set's per-instance best-of-set score. For MOSAIC the ranking is the greedy portfolio order of Section~\ref{sec:method_select}.

ReEvo additionally requires a seed heuristic function to initialize the search; its TSP seed assigns every unvisited city an identical score, its KP seed packs the first item that fits the remaining capacity, and its CVRP seed picks the feasible customer with the highest demand-to-distance ratio. The manually designed Greedy Construct baseline moves to the nearest unvisited city on TSP and packs the feasible item of highest value-to-weight density on KP.

\paragraph{Instance generation.}
\label{app:instgen_setup}
All methods evolve TSP instances of $n$ cities ($n \in \{50, 100\}$) under a fixed budget of heuristic evaluations; each setting is repeated for 3 seeds and the metrics are averaged. For the baselines, the fitness signal is the cost-ratio margin $m_{h_i, h_j}(x)$ between two reference heuristics, $h_i =$ farthest insertion and $h_j =$ nearest insertion; the budget is split equally between the two discrimination directions and the two resulting instance sets are merged for evaluation. The Ours variants run the co-evolutionary loop itself under the same budget, computing margins between their own co-evolved heuristic pairs (Section~\ref{sec:method_isa}) and returning the instances stored in the final archive.

The evolutionary algorithm (EA) baselines $(\mu{,}\lambda)$-EA~\citep{smith2011discovering}, $(\mu{+}1)$-CM-EA~\citep{bossek2019foga}, and QD-EA~\citep{bossek2022gecco} were originally proposed to evolve instances that separate pairs of inexact TSP solvers, with reference implementations in R built around the \texttt{tspgen} package\footnote{\url{https://github.com/jakobbossek/tspgen}}. We reimplement all three in Python following the \texttt{tspgen} implementation, including its full pool of 11 mutation operators, and replace the original solver pair with the two constructive reference heuristics above, so every generator competes on the same fitness signal and the same heuristic-evaluation budget as ours.
\begin{itemize}
    \item \textbf{$(\mu{,}\lambda)$-EA}~\citep{smith2011discovering} runs a generational EA with population $\mu{=}20$ (one elite per generation, $\lambda{=}19$ offspring), uniform per-city crossover, Gaussian-replace mutation with a decaying rate, and tournament selection; each run returns its best instance.
    \item \textbf{$(\mu{+}1)$-CM-EA}~\citep{bossek2019foga} is a steady-state EA with population $\mu{=}5$ and no crossover: each step mutates one random parent with a uniformly-sampled creative mutation (CM) operator from a pool of 11 (radial, structural, projection, and tube-based), and the mutant replaces the worst individual if better.
    \item \textbf{QD-EA}~\citep{bossek2022gecco} is MAP-Elites over the same feature grid, sampling parents from any filled cell and applying the same 11-operator pool.
\end{itemize}

Feature-space coverage is measured on a two-dimensional grid over two nearest-neighbor-graph features, computed on the directed 3-nearest-neighbor graph of the city set in which each city points to its three nearest other cities. This is the feature combination of \citet{bossek2022gecco} and differs from the heuristic-generation grid (Appendix~\ref{app:ours_config}):
\begin{itemize}
    \item \texttt{strong\_components\_max}: the size of the largest strongly connected component, the largest mutually reachable set of cities.
    \item \texttt{n\_weak}: the number of weakly connected components, the number of city groups that remain separated when edge directions are ignored.
\end{itemize}
Both features are integer-valued with analytically known bounds depending on $n$~\citep{HEINS2023123}, $[4, n]$ for \texttt{strong\_components\_max} and $[1, \lfloor n/4 \rfloor]$ for \texttt{n\_weak} on the 3-nearest-neighbor graph, giving a finite grid on which the coverage ratio is computed: $47 \times 12 = 564$ cells at $n{=}50$ and $97 \times 25 = 2425$ cells at $n{=}100$.

Algorithmic-performance metrics are computed against a portfolio of five traditional constructive heuristics:
\begin{itemize}
    \item \textsc{Nearest Neighbour}: moves to the closest unvisited city.
    \item \textsc{Farthest Unvisited}: moves to the farthest unvisited city.
    \item \textsc{Greedy Return}: minimizes the distance to the candidate city plus the candidate's distance to the tour's destination.
    \item \textsc{Look-Ahead NN}: minimizes the distance to the candidate plus the candidate's own nearest-unvisited distance.
    \item \textsc{Nearest-to-Visited}: selects the unvisited city closest to any already-visited city.
\end{itemize}

\paragraph{Our variants.}
We evaluate three variants of our instance generator that isolate the contributions of the LLM and of the evolutionary loop:
\begin{itemize}
    \item \textbf{Ours (Classical EA)} keeps the evolutionary loop but drives it with the crossover and mutation operators of the $(\mu{,}\lambda)$-EA baseline instead of the LLM-designed instance-generation function $g_{\text{inst}}$; tournament selection keeps the top-$N_{\text{inst}}$ instances by $m_{h_i, h_j}$.
    \item \textbf{Ours (LLM-Direct)} makes a single call to the instance-evolver LLM that produces a one-shot instance-generation function executed once to return the full set of $N_{\text{inst}}$ instances; there is no evolutionary loop and no fitness feedback.
    \item \textbf{Ours (LLM-Guided EA)} is the framework's generator (Section~\ref{sec:method_isa}): a single LLM call produces the unified generate-and-transform function $g_{\text{inst}}$, driven by the evolutionary loop of Algorithm~\ref{alg:evolve_inst}.
\end{itemize}

\subsection{LLM Configuration}
Both LLM backbones are accessed through the OpenAI API with model identifiers \texttt{gpt-4o-mini-2024-07-18} and \texttt{gpt-5-nano-2025-08-07}. MOSAIC queries every LLM role (initialization, instance evolver, reflection, crossover, and mutation) with sampling temperature $1.0$; for GPT-5-nano, whose temperature is not configurable, reasoning effort and verbosity are set to \texttt{medium}. Baselines use the sampling settings of their original implementations.

\subsection{Computing Infrastructure}
All experiments ran on a single Linux workstation (Ubuntu 22.04) with an AMD Ryzen Threadripper PRO 7985WX CPU (64 cores) and 256\,GB of RAM; LLM queries go through the OpenAI API and all heuristic evaluations are CPU-parallel. The framework is implemented in Python~3.11 with NumPy~2.2, scikit-learn~1.7 (decision trees), and joblib~1.5 (process-level parallelism); optimal reference values are computed with LKH-3~\citep{lkh3} and OR-Tools v9.12~\citep{ortools}.

\section{Evaluation Metrics}
\label{app:metrics}

\subsection{Heuristic Generation}
For each test instance $x_i$ with optimum $f^*(x_i)$, the optimality gap of heuristic $h$ is $g(h, x_i) = \big(f(h(x_i)) - f^*(x_i)\big) / f^*(x_i)$. A run of a method produces a ranked heuristic pool $h^{(1)}, h^{(2)}, \ldots$ (training-performance order for baselines, greedy portfolio order for MOSAIC); scores are averaged over a method's independent runs. At test time every heuristic call runs under a per-instance time limit of 15, 30, and 60 seconds for sizes 50, 100, and 200 on all three problems, and 120 seconds for KP $n{=}500$; a heuristic fails on an instance if it raises an error, returns an invalid solution, or exceeds this limit.

\textbf{Top-$k$.} The mean over test instances of the per-instance best gap among the first $k$ ranked heuristics, $\frac{1}{N}\sum_{i=1}^{N} \min_{j \leq k} g(h^{(j)}, x_i)$; if all $k$ heuristics fail on an instance, the next-ranked heuristics are tried in order, identically for every method.

\textbf{Oracle.} The Top-$k$ score with the run's full heuristic pool in place of the first $k$ ranked heuristics, an upper bound on any per-instance selection from that pool, including each method's single top-ranked heuristic.

\subsection{Instance Generation}
To assess the quality of instances produced by different evolutionary instance-generation methods, we evaluate each instance set along two complementary axes: \emph{feature-space coverage}, how much of the feature grid the instances occupy, and \emph{algorithmic performance}, how hard the instances are for a heuristic portfolio and how strongly they separate its heuristics.

Instances are scored on a feature grid $\mathcal{G}$ (integer features mapped to their integer value, continuous features partitioned into $R$ equal-width bins using bounds) and against a heuristic portfolio of $H$ solvers, where for each instance $i$ the optimality gap of heuristic $h$ is $g_i^h = (f_i^h - f_i^*)/f_i^*$ with $f_i^*$ the optimal objective value.

\textbf{Coverage ratio (CR).} The fraction of the feature grid filled by at least one generated instance,
\begin{equation}
    \text{CR} = \frac{|\mathcal{C}|}{|\mathcal{G}|},
\end{equation}
where $\mathcal{C}$ is the set of distinct filled cells and $|\mathcal{G}|$ is the total number of grid cells, the product of the per-axis cell counts.

\textbf{Mean optimality gap (Gap).} The average of the per-instance mean gap $\bar{g}_i = \frac{1}{H}\sum_h g_i^h$ over all instances; it measures overall instance hardness.

\textbf{Instance discrimination score (Discrim).} The mean spread of gaps across the heuristic portfolio over the $N$ generated instances,
\begin{equation}
    \bar{\Delta} = \frac{1}{N}\sum_{i=1}^{N} \Big( \max_h\, g_i^h - \min_h\, g_i^h \Big),
\end{equation}
which is high when instances are structurally selective: certain heuristics perform well while others struggle.

\section{Extended Experiments}
\label{app:extended_experiments}

\subsection{Heuristic Generation}
\label{app:cvrp}

\paragraph{Evaluation on CVRP.}
We extend the heuristic-generation comparison to the Capacitated Vehicle Routing Problem (CVRP). Heuristics are constructive, serving one customer per step under the vehicle-capacity constraint, and we compare against \textbf{ReEvo}~\citep{ye2024reevo}, \textbf{EoH-S}~\citep{eohs}, and the manually designed \textbf{Nearest Feasible} heuristic, which always serves the nearest customer whose demand fits the remaining capacity. The protocol of the main experiments is otherwise unchanged: every LLM-based method is run 9 times (3 seeds, 3 runs per seed) under the same budget of 300 generated heuristics with GPT-4o-mini and GPT-5-nano as LLM backbones, methods train at $n{=}50$, and the test sets contain 200 instances per size $n \in \{50, 100\}$, spanning uniform and clustered customer layouts and demands positively and negatively correlated with depot distance (Appendix~\ref{app:benchmarks}), with optimal values from LKH-3~\citep{lkh3}.

MOSAIC attains the lowest gap in Table~\ref{tab:cvrp_main} at every selection level, size, and LLM backbone. Relative to the best baseline, EoH-S, the Top-3 and Top-5 gaps are up to $29\%$ lower with GPT-5-nano and $7\%$ lower with GPT-4o-mini, the Oracle gap is up to $35\%$ and $10\%$ lower, and with GPT-5-nano the Top-3 gap is lower than every baseline's \emph{Oracle} gap at both sizes.

\begin{table}[t]
\centering
\small
\renewcommand{\arraystretch}{0.98}
\setlength{\extrarowheight}{-0.3pt}
\setlength{\tabcolsep}{12pt}
\resizebox{\columnwidth}{!}{%
\begin{tabular}{ll|rGr}
\bottomrule[0.5mm]
 & & \multicolumn{2}{c}{\textbf{CVRP}} \\
\hline
\textbf{Method} & \textbf{Selection} & $n{=}50$ & $n{=}100$ \\
\midrule[0.3mm]
Nearest Feasible &  & 34.17 & 31.45 \\
\hline
\multicolumn{4}{c}{LLM-based AHD: \emph{GPT-4o-mini}} \\
\hline
ReEvo         & Top-3  & 30.49 & 27.67 \\
              & Top-5  & 28.80 & 26.24 \\
              & Oracle & 26.84 & 24.82 \\
\arrayrulecolor{gray!40}\hline\arrayrulecolor{black}
EoH-S         & Top-3  & 27.88 & 26.19 \\
              & Top-5  & 26.21 & 25.13 \\
              & Oracle & 24.44 & 23.94 \\
\arrayrulecolor{gray!40}\hline\arrayrulecolor{black}
MOSAIC (Ours) & Top-3  & \textbf{25.81} & \textbf{24.23} \\
              & Top-5  & \textbf{24.29} & \textbf{23.40} \\
              & Oracle & \textbf{21.96} & \textbf{21.57} \\
\hline
\multicolumn{4}{c}{LLM-based AHD: \emph{GPT-5-nano}} \\
\hline
ReEvo         & Top-3  & 30.40 & 27.79 \\
              & Top-5  & 29.32 & 27.16 \\
              & Oracle & 27.84 & 26.00 \\
\arrayrulecolor{gray!40}\hline\arrayrulecolor{black}
EoH-S         & Top-3  & 19.42 & 18.19 \\
              & Top-5  & 18.29 & 17.46 \\
              & Oracle & 16.81 & 16.59 \\
\arrayrulecolor{gray!40}\hline\arrayrulecolor{black}
MOSAIC (Ours) & Top-3  & \textbf{13.87} & \textbf{13.32} \\
              & Top-5  & \textbf{12.95} & \textbf{12.55} \\
              & Oracle & \textbf{10.86} & \textbf{11.01} \\
\toprule[0.5mm]
\end{tabular}%
}
\caption{Heuristic generation on CVRP: mean optimality gap (\%) over runs on 200 test instances per size. LKH-3~\citep{lkh3} provides the optimal values. Best value per column and selection level within each LLM-backbone block in bold.}
\label{tab:cvrp_main}
\end{table}

\paragraph{Statistical significance testing.}
We run one-sided Mann--Whitney $U$ tests~\citep{mann1947test} over the per-run gaps of MOSAIC against LLM-based AHD methods, at the largest test size of each problem, TSP $n{=}200$ and KP $n{=}500$, with the GPT-4o-mini LLM backbone. Table~\ref{tab:significance} reports the $p$-values for the Top-3 and Oracle levels; 9 of the 10 comparisons are significant at the $0.05$ level.

\begin{table}[t]
\centering
\footnotesize
\setlength{\tabcolsep}{3pt}
\begin{tabular}{ll ccc}
\toprule
Setting & Selection & ReEvo & MCTS-AHD & EoH-S \\
\midrule
TSP $n{=}200$ & Top-3  & \textbf{.005}   & \textbf{<.001} & \textbf{.046} \\
              & Oracle & \textbf{<.001}  & \textbf{<.001} & \textbf{.018} \\
\midrule
KP $n{=}500$  & Top-3  & .057            & \textbf{.011} & --- \\
              & Oracle & \textbf{.021}   & \textbf{.006} & --- \\
\bottomrule
\end{tabular}
\caption{$p$-values of the significance test, MOSAIC vs.\ LLM-based AHD methods over per-run gaps at the largest test size (GPT-4o-mini). Bold marks significance at the $0.05$ level; EoH-S is not part of the KP comparison.}
\label{tab:significance}
\end{table}

\paragraph{Feature sensitivity.}
We test whether the results depend on the choice of the grid feature pair. We repeat the TSP experiments with GPT-4o-mini under a different pair, keeping everything else identical:
\begin{itemize}
    \item Default: \texttt{fraction\_of\_distinct\_distances} $\times$ \texttt{n\_strong} (Appendix~\ref{app:ours_config}); these are the runs of Table~\ref{tab:main_results}.
    \item Alternative: \texttt{distance\_span} $\times$ \texttt{weak\_components\_max}: the range between the largest and smallest pairwise city distance, and the size of the largest weakly connected component of the directed 3-nearest-neighbor graph.
\end{itemize}
Table~\ref{tab:feature_sensitivity} shows that the two feature pairs are statistically indistinguishable. Every difference is within one standard deviation, and the alternative pair gives slightly lower Oracle gaps and slightly higher Top-$k$ gaps. The framework's gains do not hinge on the particular feature pair.

\begin{table}[t]
\centering
\footnotesize
\setlength{\tabcolsep}{4pt}
\begin{tabular}{ll ccc}
\toprule
\textbf{Feature pair} & \textbf{Selection} & $n{=}50$ & $n{=}100$ & $n{=}200$ \\
\midrule
Default     & Top-3  & $8.62{\pm}0.70$ & $12.05{\pm}0.80$ & $14.78{\pm}0.98$ \\
            & Top-5  & $7.78{\pm}0.70$ & $11.27{\pm}0.84$ & $14.06{\pm}1.07$ \\
            & Oracle & $6.10{\pm}0.46$ & $9.71{\pm}0.61$  & $12.72{\pm}0.87$ \\
\midrule
Alternative & Top-3  & $8.87{\pm}0.54$ & $12.59{\pm}0.54$ & $15.18{\pm}0.70$ \\
            & Top-5  & $7.55{\pm}0.68$ & $11.54{\pm}0.75$ & $14.26{\pm}0.97$ \\
            & Oracle & $5.74{\pm}0.69$ & $9.48{\pm}0.75$  & $12.58{\pm}0.88$ \\
\bottomrule
\end{tabular}
\caption{Feature sensitivity on TSP with GPT-4o-mini: mean $\pm$ std over runs of the optimality gap (\%) on the test sets of Table~\ref{tab:main_results}, for the default and alternative grid feature pairs.}
\label{tab:feature_sensitivity}
\end{table}

\paragraph{Evaluation on TSPLIB instances.}
We evaluate the generated heuristics on real-world instances of the TSPLIB benchmark~\citep{reinelt1991tsplib} under the experimental setup of the main TSP experiments. Each heuristic runs three times from different starting nodes, and its three tour lengths are averaged before computing the gap to the known optimum. Table~\ref{tab:tsplib} reports the Top-5 heuristic portfolio of the best run of MOSAIC and of EoH-S~\citep{eohs}, the strongest baseline heuristic pool on TSP, each selected by the lowest mean Top-5 gap on the TSP test sets of Table~\ref{tab:main_results}, with the GPT-5-nano LLM backbone, next to the classical constructive baselines Christofides~\citep{christofides2022worst}, Greedy~\citep{brecklinghaus2015approximation}, Nearest insertion, and Nearest-greedy~\citep{rosenkrantz1977analysis}. MOSAIC attains the lowest average gap.

\begin{table}[t]
\centering
\footnotesize
\setlength{\tabcolsep}{2pt}
\resizebox{\columnwidth}{!}{%
\begin{tabular}{l rrrr rr}
\toprule
\textbf{Instance} & Christofides & Greedy & \shortstack{Nearest\\insertion} & \shortstack{Nearest-\\greedy} & EoH-S & Ours \\
\midrule
a280    & 15.37 & 26.76 & 15.38 & 32.98 & 14.33 & \textbf{14.06} \\
ch130   & 11.91 & 28.40 & 21.29 & 27.06 & 10.17 & \textbf{8.53} \\
ch150   & 9.73  & 18.61 & 22.84 & 22.20 & 6.42  & \textbf{5.75} \\
d198    & \textbf{9.74} & 23.62 & 14.25 & 25.30 & 14.87 & 14.63 \\
eil51   & 12.80 & 13.03 & 17.19 & 29.53 & 6.79  & \textbf{4.37} \\
gil262  & 13.04 & 12.46 & 22.12 & 31.16 & 11.59 & \textbf{11.33} \\
kroA100 & \textbf{9.45} & 13.70 & 21.84 & 25.34 & 11.18 & 9.56 \\
kroA150 & 10.68 & 20.24 & 18.94 & 26.52 & 10.52 & \textbf{8.98} \\
kroA200 & 12.35 & 17.83 & 22.34 & 26.39 & 11.95 & \textbf{10.58} \\
kroB150 & 14.17 & 20.25 & 21.57 & 29.97 & 11.80 & \textbf{10.82} \\
kroB200 & 11.43 & 22.24 & 23.86 & 24.69 & 11.77 & \textbf{11.30} \\
kroC100 & \textbf{9.67} & 12.94 & 23.75 & 25.02 & 10.03 & 10.22 \\
lin105  & 14.69 & 16.64 & 29.07 & 40.69 & 17.19 & \textbf{5.81} \\
lin318  & 12.93 & 18.75 & 24.42 & 28.22 & 14.98 & \textbf{11.11} \\
pr76    & 7.88  & 29.76 & 19.36 & 41.27 & \textbf{4.73} & 6.89 \\
pr136   & 7.23  & 23.67 & 11.14 & 26.16 & \textbf{6.08} & 7.23 \\
pr226   & 15.00 & 20.11 & 28.14 & 18.55 & 8.89  & \textbf{7.55} \\
pr264   & \textbf{10.97} & 14.08 & 28.69 & 19.20 & 13.96 & 12.19 \\
rat195  & 14.47 & 18.73 & 26.05 & 18.14 & \textbf{7.65} & 7.99 \\
rd100   & 12.59 & 16.99 & 20.43 & 29.82 & 12.21 & \textbf{9.28} \\
st70    & 13.20 & 10.53 & 16.95 & 18.15 & 7.61  & \textbf{6.79} \\
u159    & \textbf{11.43} & 15.64 & 23.42 & 29.18 & 13.72 & 13.93 \\
\midrule
Average Gap & 11.85 & 18.86 & 21.50 & 27.07 & 10.84 & \textbf{9.50} \\
\bottomrule
\end{tabular}%
}
\caption{Optimality gap (\%) on TSPLIB instances with the GPT-5-nano LLM backbone. Each heuristic runs 3 times from different starting nodes and its averaged tour length gives the gap to the known optimum; EoH-S and Ours report the Top-5 heuristic portfolio of each method's best run. Best value per instance in bold.}
\label{tab:tsplib}
\end{table}

\paragraph{Cost and runtime.}
Table~\ref{tab:cost_runtime} compares MOSAIC with MCTS-AHD on wall-clock time and LLM cost per run. A MOSAIC run costs about \$0.18 more: beyond the crossover and mutation queries that all methods issue, it also queries the instance-evolver and reflection LLMs, so the budget buys discriminative instances and multi-directional insights as well as specialized heuristics. A run also takes about half the wall-clock time of MCTS-AHD: within an iteration the $N_{\text{pair}}$ pair analyses of the discriminative instance generation (Section~\ref{sec:method_isa}) and the crossover and mutation queries run in parallel, whereas MCTS-AHD expands its search tree one node at a time, each expansion's LLM query and evaluation conditioned on the outcome of the previous ones.

\begin{table}[t]
\centering
\footnotesize
\setlength{\tabcolsep}{5pt}
\begin{tabular}{l cc}
\toprule
\textbf{Method} & \textbf{Time/run (min)} & \textbf{Cost/run (USD)} \\
\midrule
MCTS-AHD            & $40.6{\pm}9.2$ & $0.145{\pm}0.006$ \\
MOSAIC (Ours) & $21.9{\pm}3.9$ & $0.320{\pm}0.018$ \\
\bottomrule
\end{tabular}
\caption{Average wall-clock time and LLM cost per run on TSP with the GPT-4o-mini LLM backbone (mean $\pm$ std over runs).}
\label{tab:cost_runtime}
\end{table}

\subsection{Instance Generation}
\label{app:instgen_extended}

Table~\ref{tab:instgen_5k} repeats the baseline comparison of Table~\ref{tab:instgen_main} at both instance sizes, $n{=}50$ and $n{=}100$, under the same $5{\times}10^3$ budget. We further test whether the advantage of our generator persists with fewer heuristic evaluations. Table~\ref{tab:instgen_1k} repeats the comparison of Table~\ref{tab:instgen_5k} at one fifth of the budget ($10^3$ heuristic evaluations). The ranking is unchanged. At $n{=}50$ our generator covers $22\%$ more of the feature grid than QD-EA, and its instances are $25\%$ harder and $36\%$ more discriminating; at $n{=}100$ these margins are $41\%$, $27\%$, and $36\%$. At $n{=}50$, hardness and discrimination at $10^3$ already match their $5{\times}10^3$ levels, so the LLM-designed operators reach near-maximal discrimination on the heuristic portfolio within the first generations, and additional budget translates almost entirely into feature-space coverage.

The baselines' trade-off has a mechanistic reading: they apply blind geometric mutations against a single frozen solver pair, so budget buys either margin depth or, with a MAP-Elites grid, undirected breadth. Our generator couples pair-conditioned LLM operators that construct structurally new layouts directly, with a stream of co-evolved heuristic pairs that keeps relocating the discrimination target across the feature space, while the archive stores every validated instance and reuses it to seed later evolutions, so coverage and discrimination accumulate jointly rather than trading off.
\begin{table}[t]
\centering
\footnotesize
\setlength{\tabcolsep}{3pt}
\begin{tabular}{l ccc}
\toprule
 & \multicolumn{1}{c}{\scriptsize\textbf{Feature-Space Coverage}} & \multicolumn{2}{c}{\scriptsize\textbf{Algorithmic Performance}} \\
\cmidrule(lr){2-2} \cmidrule(l){3-4}
\textbf{Method} & \textbf{CR(\%)}$\,\uparrow$ & \textbf{Gap(\%)}$\,\uparrow$ & \textbf{Discrim(\%)}$\,\uparrow$ \\
\midrule
\multicolumn{4}{c}{$n{=}50$} \\
\midrule
$(\mu{,}\lambda)$-EA    & $3.43{\pm}0.41$  & $146.2{\pm}1.1$ & $537{\pm}9$   \\
$(\mu{+}1)$-CM-EA & $3.49{\pm}0.10$  & $147.7{\pm}1.9$ & $543{\pm}12$  \\
QD-EA                   & $25.12{\pm}1.95$ & $147.6{\pm}2.7$ & $554{\pm}14$  \\
Ours                    & $\mathbf{33.45{\pm}0.87}$ & $\mathbf{182.1{\pm}6.5}$ & $\mathbf{744{\pm}36}$ \\
\midrule
\multicolumn{4}{c}{$n{=}100$} \\
\midrule
$(\mu{,}\lambda)$-EA    & $0.89{\pm}0.02$  & $213.6{\pm}1.2$ & $846{\pm}11$  \\
$(\mu{+}1)$-CM-EA & $0.88{\pm}0.02$  & $222.0{\pm}3.9$ & $889{\pm}16$  \\
QD-EA                   & $10.96{\pm}0.35$ & $218.5{\pm}6.9$ & $874{\pm}35$  \\
Ours                    & $\mathbf{12.58{\pm}1.23}$ & $\mathbf{322.6{\pm}46.3}$ & $\mathbf{1399{\pm}234}$ \\
\bottomrule
\end{tabular}
\caption{Instance generation on TSP at a budget of $5{\times}10^3$ heuristic evaluations for both instance sizes (mean $\pm$ std over 3 seeds; GPT-5-nano LLM backbone); best value per column in bold. Ours is the framework's default generator, Ours (LLM-Guided EA) in Table~\ref{tab:instgen_main}.}
\label{tab:instgen_5k}
\end{table}

\begin{table}[t]
\centering
\footnotesize
\setlength{\tabcolsep}{3pt}
\begin{tabular}{l ccc}
\toprule
 & \multicolumn{1}{c}{\scriptsize\textbf{Feature-Space Coverage}} & \multicolumn{2}{c}{\scriptsize\textbf{Algorithmic Performance}} \\
\cmidrule(lr){2-2} \cmidrule(l){3-4}
\textbf{Method} & \textbf{CR(\%)}$\,\uparrow$ & \textbf{Gap(\%)}$\,\uparrow$ & \textbf{Discrim(\%)}$\,\uparrow$ \\
\midrule
\multicolumn{4}{c}{$n{=}50$} \\
\midrule
$(\mu{,}\lambda)$-EA    & $0.59{\pm}0.10$  & $143.7{\pm}2.7$ & $530{\pm}27$  \\
$(\mu{+}1)$-CM-EA & $0.71{\pm}0.00$  & $146.4{\pm}4.6$ & $530{\pm}8$   \\
QD-EA                   & $15.07{\pm}2.50$ & $145.3{\pm}5.0$ & $543{\pm}24$  \\
Ours                    & $\mathbf{18.32{\pm}4.55}$ & $\mathbf{181.6{\pm}20.1}$ & $\mathbf{740{\pm}110}$ \\
\midrule
\multicolumn{4}{c}{$n{=}100$} \\
\midrule
$(\mu{,}\lambda)$-EA    & $0.15{\pm}0.02$ & $215.6{\pm}5.5$ & $855{\pm}27$  \\
$(\mu{+}1)$-CM-EA & $0.16{\pm}0.00$ & $225.4{\pm}1.7$ & $902{\pm}15$  \\
QD-EA                   & $6.10{\pm}1.12$ & $217.6{\pm}8.2$ & $866{\pm}41$  \\
Ours                    & $\mathbf{8.60{\pm}1.32}$ & $\mathbf{276.7{\pm}28.1}$ & $\mathbf{1179{\pm}144}$ \\
\bottomrule
\end{tabular}
\caption{Instance generation on TSP at a budget of $10^3$ heuristic evaluations (mean $\pm$ std over 3 seeds; GPT-5-nano LLM backbone); best value per column in bold.}
\label{tab:instgen_1k}
\end{table}

\subsection{Archive and Instance-Space Visualizations}
\label{app:visualizations}

\begin{figure*}[t]
    \centering
    \includegraphics[width=\linewidth]{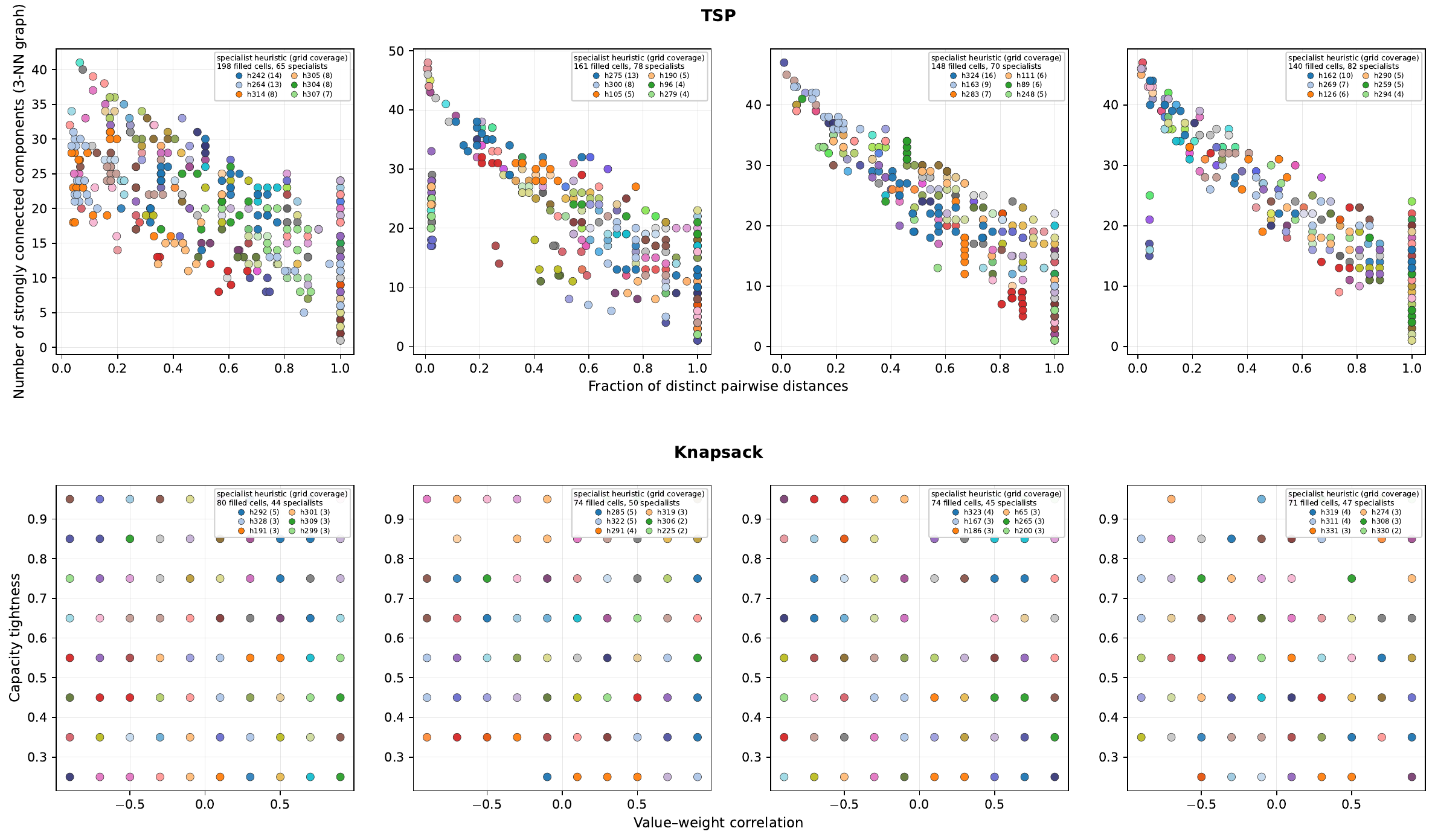}
    \caption{Final QD archives of four TSP runs (top row) and four KP runs (bottom row). In the TSP panels each dot is a stored archive instance and the axes are the fraction of distinct pairwise distances and the number of strongly connected components of the 3-NN graph. In the KP panels each dot is a filled cell and the axes are the value--weight correlation and the capacity tightness. Color marks the cell's specialist heuristic, and the legend of each panel gives its filled-cell and specialist counts.}
    \label{fig:winner_map}
\end{figure*}

\begin{figure*}[t]
    \centering
    \includegraphics[width=\linewidth]{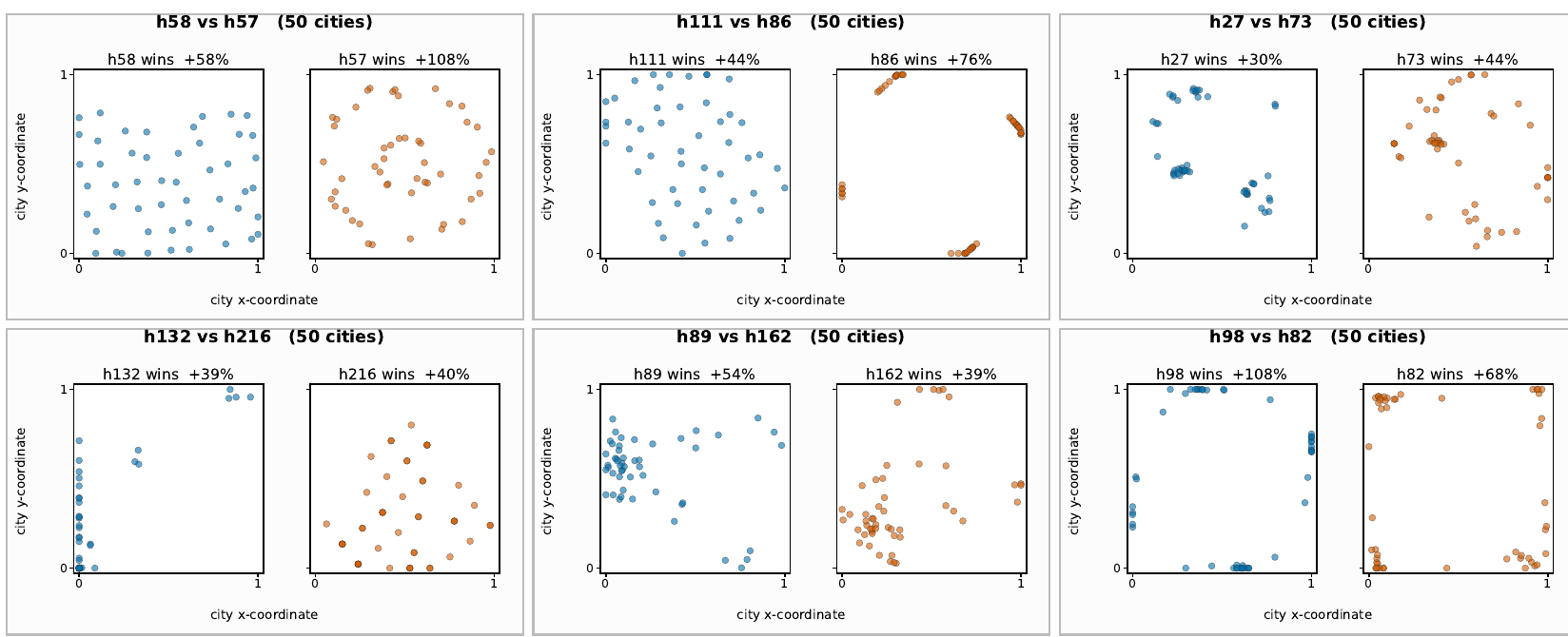}
    \caption{City layouts of discriminative instances for six heuristic pairs, evolved with 50 cities in the unit square. Each outlined box is one pair. The left panel shows the generated instance on which the first heuristic gives the shorter tour and the right panel the instance favoring the second, annotated with the winner's tour-length margin over its rival.}
    \label{fig:pair_instances}
\end{figure*}

\begin{figure*}[t]
    \centering
    \includegraphics[width=\linewidth]{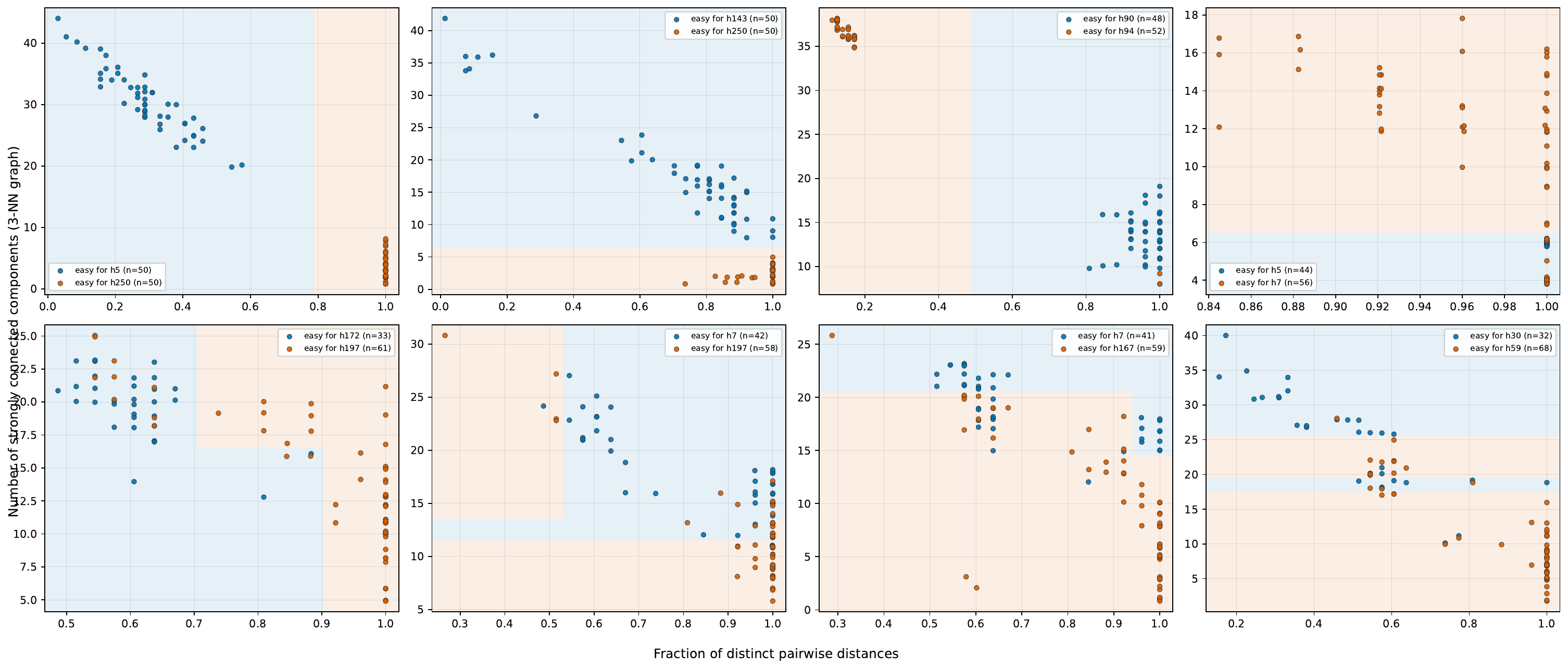}
    \caption{Decision-tree training sets for eight heuristic pairs in the run's grid feature space. Each dot is one evolved instance, colored by the heuristic that attains the shorter tour, with per-class counts in the legends. Shading marks the decision regions of the fitted depth-3 tree.}
    \label{fig:pair_scatter}
\end{figure*}

Figure~\ref{fig:winner_map} examines whether the co-evolution ends in region specialists rather than a single dominant heuristic. Each TSP panel plots the stored archive instances at their feature coordinates, so it also shows how far the co-evolved instances have expanded the filled set. Each panel shows an archive partitioned among many specialists. The TSP archives hold 140 to 198 filled cells served by 65 to 82 specialists, and the most widespread specialist owns at most 16 cells. The KP archives hold 71 to 80 filled cells served by 44 to 50 specialists and are even more finely partitioned.

Figure~\ref{fig:pair_instances} gives example outputs of the discriminative instance generation, drawn from a run to illustrate. Each box shows one heuristic pair with one evolved instance from each of its two directions, an instance on which the first heuristic wins and an instance on which the second wins. The two instances of a pair are structurally different, and the layouts vary widely across pairs, including a spiral ring, arcs along the border, well-separated clusters, a line of cities on one edge, and dense corner groups. The winner's tour-length margins range from 30\% to 108\%, so algorithmic performance differs as sharply as the structure. This is the raw signal that the per-pair decision tree converts into feature-space regions.

Figure~\ref{fig:pair_scatter} plots the evolved instances of eight analyzed pairs in the grid feature space, colored by the winning heuristic, together with the decision regions of the fitted tree, again drawn from a run to illustrate. In every pair the two classes separate along feature boundaries, in forms that range from a single threshold on one feature to interleaved regions that use both features. The correctly classified instances of each leaf are the ones that pass validation, enter the grid cells, update the specialists, and receive the leaf's insight. The figure thus shows that the reliable feature-space separation required by the grid update and the spatially-grounded insight storage indeed arises in practice.

\section{Extended Ablations}
\label{app:extended_ablations}

\begin{table*}[t]
\centering
\small
\setlength{\tabcolsep}{35pt}
\resizebox{\textwidth}{!}{%
\begin{tabular}{@{}l p{5.6cm} p{6.8cm}@{}}
\toprule
\textbf{Ablation} & \textbf{Removed or changed component} & \textbf{Replacement} \\
\midrule
\multicolumn{3}{@{}l}{\emph{QD grid archive}} \\
W/o QD grid (flat population) & The grid archive with its specialists, cube-based local evaluation, spatial insight storage, and distance-based pairing & A flat population of 20 kept by global elitism on mean training cost, with uniform-random pairs and one global insight list \\
Global offspring evaluation & Cube-based local evaluation of offspring with $\rho{=}0.4$ & Every offspring is evaluated on all filled cells ($\rho{=}1$) \\
Random parent pairs & Distance-ranked pair sampling (Section~\ref{sec:method_pairs}) & Uniform-random previously unanalyzed pairs \\
Random mutation targeting & Staleness-ranked target cube and coverage-ranked parent selection (Section~\ref{sec:method_mutation}) & Uniform-random target cube and parent \\
Static instance set & Storage of evolved instances in the archive & The archive instances stay frozen at $\mathcal{D}_{\text{init}}$, instances are still evolved and analyzed but never stored \\
W/o decision-tree filter & Decision-tree validation of instance placement and insight routing (Section~\ref{sec:method_grid_update}) & Each instance enters its feature cell with its raw per-instance winner, and each parent's insight is written to all cells that parent owns \\
\midrule
\multicolumn{3}{@{}l}{\emph{Operator mix}} \\
W/o crossover & The crossover operator & The heuristic-generation budget flows through mutation, and reflection still runs \\
W/o mutation & The mutation operator & The heuristic-generation budget flows through crossover \\
\midrule
\multicolumn{3}{@{}l}{\emph{Contrastive multi-directional feedback (Section~\ref{sec:method_grid_update})}} \\
Uni-directional feedback & Contrastive multi-directional reflection (Eq.~\ref{eq:reflection}) & Each analyzed pair yields a single better/worse hint that names the winner by mean cost on the pair's generated instances and briefly states why it wins, and this hint replaces the three insights in storage and retrieval \\
Crossover w/o insight & The hybridization insight $I_{\times}$ in the crossover prompt (Eq.~\ref{eq:crossover}) & The crossover prompt only asks to combine the two parents' mechanisms, and reflection otherwise runs unchanged \\
Mutation w/o insights & The insight bundle of the mutation prompt (Eq.~\ref{eq:mutation}) & The mutation prompt only asks for a modified heuristic, and mutations are never skipped, keeping the offspring budget unchanged \\
W/o insight persistence & Cross-iteration persistence of the cell-wise insight sets $\Xi_{\mathbf{k}}$ & All stored insights are erased at the start of every iteration, so only same-iteration insights are retrievable \\
\midrule
\multicolumn{3}{@{}l}{\emph{Insight retrieval policy (Section~\ref{sec:method_mutation})}} \\
Most-widespread insight retrieval & Rarest-first insight retrieval at mutation & The target cube's most widespread insight is retrieved first \\
Random insight retrieval & Rarest-first insight retrieval at mutation & Uniform-random pick among the target cube's stored insights \\
\midrule
\multicolumn{3}{@{}l}{\emph{Portfolio selection (evaluation-only)}} \\
Top-$k$ by mean validation cost & Greedy complementary selection (Eq.~\ref{eq:portfolio}) & The $k$ heuristics with the lowest mean cost on $\mathcal{D}_{\text{val}}$, on the same heuristic pools as the default configuration \\
\midrule
\multicolumn{3}{@{}l}{\emph{Parameter sweeps}} \\
$\rho \in \{0.1, 0.25\}$ & Cube fraction of each axis (default 0.4) & Global offspring evaluation provides the $\rho{=}1$ endpoint \\
$R \in \{3, 25\}$ & Bin count for continuous features (default 10) & --- \\
$L \in \{1, 4\}$ & Max insights per mutation prompt (default 2) & Mutation w/o insights provides the $L{=}0$ endpoint \\
$N_{\text{pair}} \in \{5, 20\}$ & Pairs analyzed per iteration (default 10) & The heuristic-generation budget stays fixed, so the iteration count adapts \\
\bottomrule
\end{tabular}%
}
\caption{Ablation definitions. Each ablation removes or changes exactly one component of the default configuration and keeps everything else identical.}
\label{tab:ablation_defs}
\end{table*}

\begin{table*}[t]
\centering
\footnotesize
\setlength{\tabcolsep}{22pt}
\resizebox{\textwidth}{!}{%
\begin{tabular}{l rrr rrr r}
\toprule
& \multicolumn{3}{c}{$n{=}50$} & \multicolumn{3}{c}{$n{=}100$} & \\
\cmidrule(lr){2-4}\cmidrule(lr){5-7}
\textbf{Ablation} & Top-3 & Top-5 & Oracle & Top-3 & Top-5 & Oracle & \textbf{AVG} \\
\midrule
MOSAIC (default)               & \textbf{9.02} & \textbf{8.31} & \textbf{6.69} & \textbf{12.12} & \textbf{11.38} & \textbf{9.89} & \textbf{9.57} \\
\arrayrulecolor{gray!40}\midrule\arrayrulecolor{black}
\multicolumn{8}{l}{\emph{QD grid archive}} \\
\quad W/o QD grid (flat population)     & 11.58 & 11.49 & 11.49 & 14.81 & 14.66 & 14.66 & 13.12 \\
\quad Global offspring evaluation       & 9.93  & 9.08  & 7.67  & 13.68 & 12.74 & 11.25 & 10.73 \\
\quad Random parent pairs               & 9.56  & 8.77  & 7.66  & 13.31 & 12.62 & 11.16 & 10.51 \\
\quad Random mutation targeting         & 11.67 & 10.87 & 9.01  & 15.59 & 14.63 & 12.95 & 12.45 \\
\quad Static instance set               & 9.46  & 8.62  & 8.07  & 12.55 & 11.78 & 10.95 & 10.24 \\
\quad W/o decision-tree filter          & 9.59  & 8.73  & 7.29  & 13.18 & 12.38 & 10.95 & 10.35 \\
\arrayrulecolor{gray!40}\midrule\arrayrulecolor{black}
\multicolumn{8}{l}{\emph{Operator mix}} \\
\quad W/o crossover                     & 9.47  & 8.60  & 6.76  & 13.02 & 12.14 & 10.09 & 10.01 \\
\quad W/o mutation                      & 13.07 & 13.07 & 13.07 & 17.48 & 17.48 & 17.48 & 15.28 \\
\arrayrulecolor{gray!40}\midrule\arrayrulecolor{black}
\multicolumn{8}{l}{\emph{Contrastive multi-directional feedback}} \\
\quad Uni-directional feedback          & 9.86  & 8.92  & 7.10  & 13.59 & 12.66 & 10.65 & 10.46 \\
\quad Crossover w/o insight             & 11.66 & 10.79 & 9.00  & 15.41 & 14.40 & 12.46 & 12.29 \\
\quad Mutation w/o insights             & 10.63 & 10.17 & 9.42  & 14.26 & 13.79 & 13.18 & 11.91 \\
\quad W/o insight persistence           & 10.13 & 9.51  & 8.42  & 13.87 & 13.22 & 12.18 & 11.22 \\
\arrayrulecolor{gray!40}\midrule\arrayrulecolor{black}
\multicolumn{8}{l}{\emph{Insight retrieval policy}} \\
\quad Most-widespread insight retrieval & 9.37  & 8.48  & 7.25  & 12.71 & 11.98 & 10.68 & 10.08 \\
\quad Random insight retrieval          & 10.02 & 9.14  & 7.80  & 13.46 & 12.49 & 11.17 & 10.68 \\
\arrayrulecolor{gray!40}\midrule\arrayrulecolor{black}
\multicolumn{8}{l}{\emph{Portfolio selection}} \\
\quad Top-$k$ by mean validation cost   & 10.16 & 9.19  & 6.69 & 12.16 & 11.69 & 9.89 & 9.96 \\
\arrayrulecolor{gray!40}\midrule\arrayrulecolor{black}
\multicolumn{8}{l}{\emph{Parameter sweeps}} \\
\quad $\rho{=}0.1$                      & 9.77  & 8.92  & 7.63  & 13.99 & 13.26 & 11.09 & 10.78 \\
\quad $\rho{=}0.25$                     & 9.98  & 9.01  & 7.26  & 13.35 & 12.34 & 10.67 & 10.43 \\
\quad $R{=}3$                           & 11.21 & 10.43 & 9.91  & 14.64 & 14.20 & 13.72 & 12.35 \\
\quad $R{=}25$                          & 9.93  & 9.10  & 7.52  & 13.32 & 12.62 & 11.06 & 10.59 \\
\quad $L{=}1$                           & 10.90 & 10.09 & 8.26  & 14.41 & 13.70 & 11.94 & 11.55 \\
\quad $L{=}4$                           & 9.97  & 9.05  & 7.76  & 13.76 & 12.94 & 11.58 & 10.84 \\
\quad $N_{\text{pair}}{=}5$             & 9.56  & 8.72  & 7.24  & 12.92 & 12.09 & 10.98 & 10.25 \\
\quad $N_{\text{pair}}{=}20$            & 9.80  & 8.94  & 7.24  & 13.86 & 12.74 & 10.81 & 10.56 \\
\bottomrule
\end{tabular}%
}
\caption{Full ablation results on TSP with the GPT-4o-mini LLM backbone: mean optimality gap (\%) at each selection level and size under the protocol of Table~\ref{tab:ablation_main}, with AVG the mean of the six values; lower is better and the default configuration is in bold. The top-$k$-by-mean ablation shares its Oracle values with the default configuration because it reuses the same heuristic pools.}
\label{tab:ablation_full}
\end{table*}

\subsection{Protocol}
Every ablation runs the experimental backbone of the main TSP experiments unchanged and modifies exactly one component. The backbone uses the GPT-4o-mini LLM backbone, the budget of 300 generated heuristics, training at $n{=}50$, the grid features and hyperparameters of Appendix~\ref{app:ours_config}, and the greedy portfolio extraction of Section~\ref{sec:method_select}. Each ablation is run 6 times with 3 training-dataset seeds and 2 runs per seed. Runs are evaluated at sizes 50 and 100 on the first 100 test instances of each size, and AVG denotes the mean over the three selection levels, Top-3, Top-5, and Oracle, at both sizes. A run whose archive collapses to a single heuristic terminates early, because parent pair selection and the pair analysis require two distinct heuristics, so no further discriminative instances, insights, or offspring can be produced.

\subsection{Ablation Definitions}
Table~\ref{tab:ablation_defs} lists every ablation, the removed or changed component, and the naive replacement. All ablations except the portfolio selection are training-time interventions with their own runs. The portfolio-selection ablation is evaluation-only and reuses the heuristic pools of the default configuration, so only the selection rule differs. The parameter sweeps vary one hyperparameter of Table~\ref{tab:hyperparams} at a time.

\subsection{Results}
Table~\ref{tab:ablation_full} reports each selection level and size behind the AVG scores of Table~\ref{tab:ablation_main}, together with the portfolio-selection ablation and the parameter sweeps. The analysis complements Section~\ref{sec:exp_ablation} and is limited to results not covered there.

\paragraph{Early termination under collapse.}
All runs of the w/o-mutation ablation collapse to a single heuristic before exhausting the heuristic-generation budget, and their partial heuristic pools hold two to three heuristics, so their Top-3, Top-5, and Oracle levels coincide by construction. The w/o-QD-grid ablation caps its heuristic pool at its population size of 20 and the surviving heuristics are near-duplicates, so its Top-5 matches its Oracle at both sizes. Removing crossover keeps heuristic pools comparable to the default configuration in size, so it shrinks portfolio complementarity rather than the heuristic pool itself.

\paragraph{QD grid archive.}
Evaluating offspring on all filled cells worsens AVG by 12\%, because a region specialist is then dismissed for losing in distant, irrelevant regions. Random parent pairs worsen AVG by 10\%, since similar parents yield weak margins, shallow decision trees, and generic insights, weakening the discriminative signal of the pair analysis. Random mutation targeting worsens AVG by 30\%, since stale regions then no longer receive targeted updates.

\paragraph{Contrastive multi-directional feedback.}
Removing insight persistence erases the accumulated regional knowledge at every iteration, so each region must relearn what wins there, worsening AVG by 17\%, more than the 9\% cost of the uni-directional feedback. Persistence thus contributes more than feedback content, consistent with the role of persistent insights as a region-aware prior (Appendix~\ref{sec:appendix_bayesian}).

\paragraph{Portfolio selection.}
Ranking each run's heuristic pool by mean cost on $\mathcal{D}_{\text{val}}$ instead of the greedy complementary selection of Eq.~\ref{eq:portfolio} leaves the Oracle unchanged by construction and worsens AVG by 4\%, because the mean-cost ranking picks strong but correlated heuristics whose per-instance wins overlap, while the greedy selection credits a heuristic only for instances it improves. The Top-$k$ gains of Table~\ref{tab:main_results} thus stem mainly from the heuristic pool rather than the selection rule.

\paragraph{Parameter sweeps.}
The sweeps support the default configuration. Evaluation locality is best at the default $\rho{=}0.4$. Tighter cubes worsen AVG by 13\% at $\rho{=}0.1$ and 9\% at $\rho{=}0.25$, since very local evaluation keeps noisy winners of very small regions, and the global endpoint $\rho{=}1$, given by the global-offspring-evaluation ablation, worsens it by 12\%. The bin count $R$ applies only to the continuous feature axis, because the integer axis bins at its natural unit (Appendix~\ref{sec:design_decisions}). A coarse grid with $R{=}3$ worsens AVG by 29\%, since structurally distinct regimes then share cells and are served by a single specialist, while $R{=}25$ worsens it by 11\% by spreading the stored instances over many sparsely filled cells. The number of retrieved insights is best at the default $L{=}2$. A single retrieved insight worsens AVG by 21\% and four insights by 13\%, so the default balances insufficient guidance against overloaded mutation prompts, with mutation w/o insights as the $L{=}0$ endpoint. Varying the pairs analyzed per iteration at the fixed heuristic-generation budget trades breadth within an iteration against the number of iterations, and $N_{\text{pair}}{=}5$ and $N_{\text{pair}}{=}20$ worsen AVG by 7\% and 10\%.

\section{Co-Evolutionary Objective: Detailed Decomposition}
\label{sec:appendix_objective_detail}

This section gives the explicit form of the instance-generation sub-objective referenced from Section~\ref{sec:prelim_objective}, and re-derives the minimax game from the two sub-objectives.

\paragraph{Instance-generation objective (explicit form).}
Let the current heuristic portfolio be $\mathcal{H}_{\mathcal{G}}$ and let $\bigcup_{\mathbf{k}} X_{\mathbf{k}}$ denote the union of all stored instances. Combining mean discrimination with the coverage incentive yields
\begin{equation}
    \max_{\{X_{\mathbf{k}}\}} \;\; \underbrace{\frac{1}{|\bigcup_{\mathbf{k}} X_{\mathbf{k}}|} \sum_{x \in \bigcup_{\mathbf{k}} X_{\mathbf{k}}} \Delta(x)}_{\text{algorithmic discrimination}} \;+\; \lambda \, \underbrace{\frac{|\mathcal{C}|}{|\mathcal{G}|}}_{\text{feature-space coverage}},
    \label{eq:instance_obj_explicit}
\end{equation}
with $\lambda > 0$. The first term rewards instances that expose performance differences across the heuristic portfolio; the second rewards populating previously empty cells. The coverage term must be optimized by search rather than direct construction: structural features are typically emergent properties of an instance (e.g., nearest-neighbor-graph statistics), so $\boldsymbol{\varphi}$ cannot be inverted to produce an instance for a prescribed cell, and unfilled regions must instead be discovered by evolving instances in the feature space~\citep{SMITHMILES2015evolveinstance}. Eq.~\ref{eq:discrimination} defines $\Delta(x)$. We operationalize this objective pairwise during heuristic discovery (Section~\ref{sec:method_isa}) by replacing $\Delta(x)$ with the cost-ratio margin $m_{h_i, h_j}(x)$ between selected heuristic pairs, which is a tractable proxy for portfolio-level discrimination.

\paragraph{From two sub-objectives to a minimax game.}
The heuristic-generation sub-objective (Eq.~\ref{eq:heuristic_obj}) seeks specialists that maximize total performance across cells; the instance-generation sub-objective (Eq.~\ref{eq:instance_obj_explicit}) seeks instances that simultaneously (i) make the per-cell average $P(h_{\mathbf{k}}; X_{\mathbf{k}})$ small under the current $\{h_{\mathbf{k}}\}$ (since high $\Delta$ means at least one $h$ does poorly on $x$, dragging down the mean of any incumbent that includes $x$ in $X_{\mathbf{k}}$) and (ii) populate previously empty cells. Letting the instance player minimize $\sum_{\mathbf{k}} P(h_{\mathbf{k}}; X_{\mathbf{k}})$ while being credited $\lambda |\mathcal{C}|/|\mathcal{G}|$ for coverage---and letting the heuristic player maximize the same sum---recovers the minimax in Eq.~\ref{eq:objective}. The coverage term is sign-flipped between the two views because it is a \emph{joint} reward: increasing $|\mathcal{C}|$ benefits both players (more cells = more useful tasks for the heuristic player; harder coverage of the space for the instance player). The reduction is one-directional: instances uniformly hard for every heuristic also minimize the sum while discriminating nothing, which is why Section~\ref{sec:method_isa} operationalizes the inner minimization through pairwise cost-ratio margins that reward separation directly rather than hardness alone.

\section{Key Design Decisions}
\label{sec:design_decisions}

\paragraph{Type-aware feature discretization.}
Instance features span fundamentally different types. Integer features (e.g., number of connected components) have natural integer values where each step corresponds to a meaningfully different instance structure. Continuous features (e.g., distance variance) require binning into a configurable number of intervals. The type-aware discretization in Eq.~\ref{eq:discretization} assigns each feature its own resolution, producing cells that correspond to meaningful instance categories rather than arbitrary numerical intervals: an integer feature gets one cell per value in its bounded range, and a continuous feature is split into $R_i$ equal-width bins.

\paragraph{Contrastive multi-directional vs.\ scalar feedback.}
Existing LLM-based AHD methods provide only scalar good/bad verbal feedback: ``this heuristic is better/worse.'' Our contrastive multi-directional reflection produces three insights that isolate the distinct algorithmic mechanism behind each parent, one improvement per mechanism, and a hybridization of the two. The insights are deliberately feature-free and self-contained; the \emph{where} comes from storage, since each insight is written only into the decision-tree-validated cells its heuristic wins, so any later mutation can use it without seeing the originating pair. This transforms feedback from a single uninformative signal into structured, spatially-anchored guidance that the LLM can directly translate into targeted modifications for specific instance regions.

\paragraph{Mechanism-diversity pressure.}
Two forces push LLM-driven heuristic search toward a monoculture: prompt-level complexity constraints that confine candidates to single-pass scoring rules, excluding mechanisms such as look-ahead or remaining-cost estimation, and popularity-based retrieval, in which the most widespread insight is fed back and re-generates the mechanism that produced it. MOSAIC counters both. The heuristic-generation prompts permit heavier per-candidate computation under an explicit efficiency guard (vectorized evaluation, restricted to the most promising candidates), the reflection prompt asks for mechanisms absent from the current pair, and mutation retrieves the rarest insights in its target region rather than the most-covered one. No complete algorithm is ever prescribed in the prompts; the search still has to discover useful mechanisms itself.

\paragraph{Staleness-driven exploration.}
The mutation phase uses staleness scores ($\sigma(\mathbf{c}_{\mathbf{k}}) = (\nu_{\mathbf{k}}+1)^{-1}$) to direct exploration toward least-updated regions, addressing a common failure mode in QD algorithms where certain regions become stagnant. By tracking update frequency and biasing mutation toward low-update regions, MOSAIC maintains continuous improvement pressure across the entire archive.

\enlargethispage{2\baselineskip}
\paragraph{Cube-based local evaluation.}
Rather than evaluating each new heuristic on the entire grid, we evaluate on local cubes. This focuses computation on the region where the new heuristic is most likely to contribute, and prevents a region-specialized heuristic from being dismissed for poor performance in distant, irrelevant regions. The cube-expansion rule ($\geq T$ filled cells) ensures statistically meaningful evaluation even when the local region is sparsely populated. In mutation, the same cube also scopes insight retrieval (Section~\ref{sec:method_mutation}), so $\rho$ jointly sets the locality of the evaluation context and of the retrieved insights; the $\rho$ sweep of Appendix~\ref{app:extended_ablations} spans this trade-off empirically.

\section{Bayesian Interpretation}
\label{sec:appendix_bayesian}

The interplay between insight-conditioned heuristic generation (Section~\ref{sec:method_operators}) and the cell-level replacement rule (Section~\ref{sec:method_grid_update}) admits a natural Bayesian reading: each cell seeks the posterior-maximizing specialist,
\begin{equation}
    h_{\mathbf{k}}^{*} = \arg\max_{h}\; p(h \mid X_{\mathbf{k}}, \Xi_{\mathbf{k}}) = \arg\max_{h}\; p(X_{\mathbf{k}} \mid h)\, p(h \mid \Xi_{\mathbf{k}}),
    \label{eq:bayesian}
\end{equation}
where the prior $p(h \mid \Xi_{\mathbf{k}})$ is a \emph{dynamic, region-aware} distribution distilled from accumulated insights and the likelihood $p(X_{\mathbf{k}} \mid h)$ favors heuristics with higher local performance. The two factors are operationalized separately: the prior by candidate-heuristic generation, since crossover and mutation sample heuristics from an LLM conditioned on retrieved insights, and the likelihood by the cell-level replacement rule, which keeps whichever candidate achieves higher $P(h; X_{\mathbf{k}})$. Proposing from the prior and accepting by likelihood comparison approximately maximizes the posterior in Eq.~\ref{eq:bayesian}.

\paragraph{Discriminative instances as active experimental design.}
When comparing two heuristics on an instance $x$, the cost-ratio margin $m_{h_i, h_j}(x)$ corresponds to the Bayes factor between their likelihoods. Evolving instances that maximize $m_{h_i, h_j}(x)$ (Section~\ref{sec:method_isa}) is therefore equivalent to active experimental design: we seek data with maximal power to discriminate the candidate heuristics, which is what a Bayes-factor-driven experimenter would acquire.

\enlargethispage{2\baselineskip}
\paragraph{Persistent insights as a learned prior.}
Because $\Xi_{\mathbf{k}}$ persists across iterations and is updated by every contrastive reflection whose validated instances reach cell $\mathbf{k}$ (Section~\ref{sec:method_grid_update}), the prior $p(h \mid \Xi_{\mathbf{k}})$ accumulates evidence about which structural strategies tend to win in region $\mathbf{k}$. This is the role of a hierarchical Bayesian prior shared across tasks in the same regime, and it distinguishes our reflection mechanism from the ephemeral scalar verbal feedback used in prior LLM-based AHD: scalar reflections cannot accumulate, so they correspond to using a flat prior every iteration. Under this interpretation, MOSAIC performs sequential posterior updates in which data (likelihood) and reflections (prior) jointly determine the specialist in each region of the instance space.

\onecolumn
\section{Algorithm Pseudocodes}
\label{sec:pseudocodes}

We formalize MOSAIC through ten algorithms organized into five groups: \textbf{(1)}~the main evolution loop and pair selection, \textbf{(2)}~instance-space analysis and instance evolution, \textbf{(3)}~grid update and insight generation, \textbf{(4)}~crossover, mutation, and local evaluation, and \textbf{(5)}~test-time portfolio extraction. Table~\ref{tab:notation} summarizes the notation used in the pseudocode, Table~\ref{tab:hyperparams} (Appendix~\ref{app:experimental_details}) lists the hyperparameter values, and Table~\ref{tab:helpers} lists helper functions used across the pseudocode.

\begin{table}[H]
\centering
\small
\setlength{\tabcolsep}{4pt}
\begin{tabular}{@{}lp{5.6cm} lp{5.6cm}@{}}
\toprule
\textbf{Symbol} & \textbf{Description} & \textbf{Symbol} & \textbf{Description} \\
\midrule
$\mathcal{X},\mathcal{S}$ & Instance / solution spaces & $\mathcal{B}(\bar{\mathbf{k}}, \rho)$ & Cube, axis fraction $\rho$, center $\bar{\mathbf{k}}$ \\
$h \in \mathcal{H}$ & Heuristic $h: \mathcal{X} \to \mathcal{S}$ & $\mathcal{T}_{ij}$ & Decision tree for pair $(h_i, h_j)$ \\
$f(s)$ & Cost function & $\mathcal{R}_{h_i}, \mathcal{R}_{h_j}$ & Per-class rule sets from $\mathcal{T}_{ij}$ \\
$P(h, x)$ & $-f(h(x))$ (higher is better) & $m_{h_i, h_j}(x)$ & Cost-ratio margin (Eq.~\ref{eq:margin}) \\
$P(h; \mathcal{D})$ & Mean perf.\ on $\mathcal{D}$ (Eq.~\ref{eq:performance}) & $\sigma(\mathbf{c}_{\mathbf{k}})$ & Staleness, $(\nu_{\mathbf{k}}+1)^{-1}$ \\
$\boldsymbol{\varphi}(x) \in \mathbb{R}^d$ & Feature vector of $x$ & $\mathcal{H}_{\mathcal{G}}$ & Unique heuristics in $\mathcal{G}$ \\
$\mathcal{G}$ & QD grid archive & $\boldsymbol{\pi}_{\text{reflection}}$ & Reflection LLM \\
$\mathbf{c}_{\mathbf{k}}$ & Grid cell at coord.\ $\mathbf{k}$ & $\boldsymbol{\pi}_{\text{crossover}}$ & Crossover LLM \\
$h_{\mathbf{k}}, X_{\mathbf{k}}, \Xi_{\mathbf{k}}$ & Specialist, instances, insights at $\mathbf{k}$ & $\boldsymbol{\pi}_{\text{mutation}}$ & Mutation LLM \\
$\nu_{\mathbf{k}}$ & Update counter of cell $\mathbf{k}$ & $\boldsymbol{\pi}_{\text{instance-evolver}}$ & Instance-evolver LLM \\
$\kappa(\cdot)$ & Coord.\ map (Eq.~\ref{eq:discretization}) & $X_{\text{seed}}$ & Seed instances for instance evolution \\
$\mathcal{C}(t)$ & Filled cells at iter.\ $t$ & $g_{\text{inst}}$ & Pair generate-and-transform operator (Eq.~\ref{eq:inst_gen_llm}) \\
$\mathcal{C}(h)$ & Cells with $h_{\mathbf{k}} = h$ & $A_k$ & Top-$k$ test-time portfolio (Eq.~\ref{eq:portfolio}) \\
$n(h) = |\mathcal{C}(h)|$ & Grid coverage of $h$ & $\mathcal{D}_{\text{init}}$ & Initial seed instances \\
$\bar{\boldsymbol{\varphi}}(h)$ & Centroid of $h$ (Eq.~\ref{eq:centroid}) & $\mathcal{D}_{\text{val}}$ & Validation instances (portfolio extraction) \\
\bottomrule
\end{tabular}
\caption{Summary of the notation used in the pseudocode.}
\label{tab:notation}
\end{table}

\begin{table}[H]
\centering
\small
\begin{tabular}{@{}lp{10cm}@{}}
\toprule
\textbf{Function} & \textbf{Description} \\
\midrule
$\Call{InitializeHeuristics}{\,}$ & Prompt the LLM with problem description and function signature to generate $N_{\text{init}}$ initial heuristic candidates. \\
$\Call{BuildCube}{\mathbf{k}, \rho, T}$ & Construct cube $\mathcal{B}(\mathbf{k}, \rho)$ centered at $\mathbf{k}$, spanning a fraction $\rho$ of each axis's resolution; expand to $\geq T$ filled cells by adding nearest cells (Euclidean distance in normalized feature space). \\
$\Call{GetStoredInstances}{\mathcal{G}, h}$ & Retrieve instances from cells where $h$ is the current specialist, i.e., $\bigcup_{\mathbf{k} \in \mathcal{C}(h)} X_{\mathbf{k}}$. \\
$\Call{ExtractFeatures}{x}$ & Compute the $d$-dimensional feature vector $\boldsymbol{\varphi}(x) \in \mathbb{R}^d$ for instance $x$. \\
$\Call{TrainDecisionTree}{\cdot, D_{\max}, \text{min\_leaf}}$ & Train a CART classifier with max depth $D_{\max}$ and minimum leaf size on $\{(\boldsymbol{\varphi}(x), y(x))\}$. \\
$\Call{ExtractRules}{\mathcal{T}}$ & Extract per-class rule sets $\mathcal{R}_{h_i}, \mathcal{R}_{h_j}$ and per-leaf instance assignments from $\mathcal{T}$. \\
\bottomrule
\end{tabular}
\caption{Helper functions used in pseudocode.}
\label{tab:helpers}
\end{table} The pseudocode follows the heuristic-generation mode; in the instance-generation experiments the same loop is used with two differences: the outer loop terminates on a budget of heuristic-on-instance evaluations rather than on the number of generated heuristics, and every instance evaluated during pair analysis is additionally recorded with its objective values to form the generator's output set.

\subsection{Main Evolution Loop and Pair Selection}
\label{sec:pseudo_main}

Algorithm~\ref{alg:main} (MOSAIC) presents the top-level procedure. After seeding the grid archive with $\mathcal{D}_{\text{init}}$ and bootstrapping it with $N_{\text{init}}$ LLM-generated heuristics, the main loop iterates through five phases per iteration (steps 1--6 of Figure~\ref{fig:overview}, steps 2 and 3 forming the instance-space analysis): parent pair selection, instance-space analysis (discriminative instance generation and decision-tree analysis), grid update with contrastive multi-directional reflection, pair crossover, and region-aware mutation. Each iteration increments the iteration counter $t$ (used for insight tagging) and the evaluation counter $e$ (used for budget control).

Algorithm~\ref{alg:select_pairs} (\textsc{SelectPairs}) details how $N_{\text{pair}}$ heuristic pairs are selected for comparative analysis. The key idea is to pair heuristics that occupy \emph{distant} regions of the feature space, as such pairs are most likely to exhibit complementary strengths and yield maximally discriminative instances. Pairs are ranked by centroid distance and sampled via inverse-rank probability.

\enlargethispage{3\baselineskip}
\begin{algorithm}[H]
\caption{MOSAIC: Map Of Specialists via Adversarial Instance Co-evolution}
\label{alg:main}
\small
\begin{algorithmic}[1]
\Require Initial dataset $\mathcal{D}_{\text{init}}$, LLMs $\boldsymbol{\pi}_{\text{reflection}}, \boldsymbol{\pi}_{\text{crossover}}, \boldsymbol{\pi}_{\text{mutation}}$, max evaluations $E_{\max}$, pairs per iteration $N_{\text{pair}}$, number of initial heuristics $N_{\text{init}}$, max instances per cell $M$
\Ensure Grid archive $\mathcal{G}$ with region-specialized heuristics
\Statex \textbf{--- Seed grid from initial dataset $\mathcal{D}_{\text{init}}$ ---}
\For{each $x \in \mathcal{D}_{\text{init}}$}
    \State $\mathbf{k} \gets \kappa(\boldsymbol{\varphi}(x))$ \Comment{Map instance to its grid cell (Eq.~\ref{eq:discretization})}
    \If{$|X_{\mathbf{k}}| < M$}
        \State $X_{\mathbf{k}}.\text{add}(x)$ \Comment{Store representative instance in cell}
    \EndIf
\EndFor
\State $\{h_1, \ldots, h_{N_{\text{init}}}\} \gets \Call{InitializeHeuristics}{\,}$ \Comment{Generate $N_{\text{init}}$ initial heuristics via LLM}
\For{each $h \in \{h_1, \ldots, h_{N_{\text{init}}}\}$} \Comment{Bootstrap archive: assign best heuristic per cell}
    \For{each $\mathbf{c}_{\mathbf{k}} \in \mathcal{C}$}
        \State $P(h; X_{\mathbf{k}}) \gets \frac{1}{|X_{\mathbf{k}}|} \sum_{x \in X_{\mathbf{k}}} P(h, x)$
        \If{$h_{\mathbf{k}} = \varnothing$ \textbf{or} $P(h; X_{\mathbf{k}}) > P(h_{\mathbf{k}}; X_{\mathbf{k}})$} \Comment{Replace if better}
            \State $h_{\mathbf{k}} \gets h$;\; $\nu_{\mathbf{k}} \gets \nu_{\mathbf{k}} + 1$
        \EndIf
    \EndFor
\EndFor
\State $e \gets N_{\text{init}}$;\; $t \gets 0$ \Comment{Evaluation and iteration counters; initial candidates count toward the budget}
\Statex \textbf{--- Main co-evolutionary loop ---}
\While{$e < E_{\max}$}
    \State $t \gets t + 1$
    \Statex \quad \textit{// Phase 1: Select $N_{\text{pair}}$ heuristic pairs for comparative analysis (Section~\ref{sec:method_pairs})}
    \State $\mathcal{Q} \gets \Call{SelectPairs}{\mathcal{G}, N_{\text{pair}}, N_{\text{cand}}}$ \Comment{Inverse-rank sampling on centroid distance (Alg.~\ref{alg:select_pairs})}
    \Statex \quad \textit{// Phase 2: Instance space analysis for each pair (Section~\ref{sec:method_isa})}
    \For{each $(h_i, h_j) \in \mathcal{Q}$ \textbf{in parallel}}
        \State $\textit{analysis}_{ij} \gets \Call{InstanceSpaceAnalysis}{h_i, h_j, \mathcal{G}}$ \Comment{Evolve instances, train tree (Alg.~\ref{alg:isa})}
    \EndFor
    \Statex \quad \textit{// Phase 3: Update grid with new instances and generate contrastive insights (Section~\ref{sec:method_grid_update})}
    \For{each $(h_i, h_j) \in \mathcal{Q}$}
        \State $\mathcal{G}, I_{\times}^{ij} \gets \Call{UpdateGridAndGenerateInsights}{\mathcal{G}, h_i, h_j, \textit{analysis}_{ij}, \boldsymbol{\pi}_{\text{reflection}}, t}$ \Comment{Alg.~\ref{alg:grid_update}}
    \EndFor
    \Statex \quad \textit{// Phase 4: LLM-guided crossover for each pair (Section~\ref{sec:method_crossover})}
    \For{each $(h_i, h_j) \in \mathcal{Q}$}
        \State $h' \gets \Call{LLMCrossover}{\boldsymbol{\pi}_{\text{crossover}}, h_i, h_j, I_{\times}^{ij}}$ \Comment{Generate offspring and evaluate it on the parent-centered cube (Alg.~\ref{alg:crossover})}
    \EndFor
    \Statex \quad \textit{// Phase 5: Region-aware mutation, one per pair (Section~\ref{sec:method_mutation})}
    \For{$m = 1, \ldots, N_{\text{pair}}$} \Comment{$N_{\text{pair}}$ mutations matching the number of pairs}
        \State $h', \mathcal{B} \gets \Call{InsightDrivenMutation}{\mathcal{G}, \boldsymbol{\pi}_{\text{mutation}}, N_{\text{cand}}, C, L, t}$ \Comment{Mutate toward stale region (Alg.~\ref{alg:mutation})}
        \State $\Call{EvaluateOnCube}{\mathcal{G}, h', \mathcal{B}}$ \Comment{Evaluate and replace if better (Alg.~\ref{alg:eval_cube})}
    \EndFor
\EndWhile
\State \Return $\mathcal{G}$
\end{algorithmic}
\end{algorithm}

\begin{algorithm}[H]
\caption{\textsc{SelectPairs}: Distance-Weighted Heuristic Pair Selection}
\label{alg:select_pairs}
\small
\begin{algorithmic}[1]
\Require Grid archive $\mathcal{G}$, number of pairs $N_{\text{pair}}$, truncation size $N_{\text{cand}}$
\Ensure Set of $N_{\text{pair}}$ heuristic pairs $\mathcal{Q}$
\State $\mathcal{H}_{\mathcal{G}} \gets \{h : |\mathcal{C}(h)| > 0\}$ \Comment{Heuristics with grid presence}
\For{each $h \in \mathcal{H}_{\mathcal{G}}$} \Comment{Compute centroid of each heuristic's cells}
    \State $\bar{\boldsymbol{\varphi}}(h) \gets \frac{1}{|\mathcal{C}(h)|}\sum_{\mathbf{k} \in \mathcal{C}(h)} \tilde{\mathbf{k}}$ \Comment{Normalized centroid (Eq.~\ref{eq:centroid})}
\EndFor
\State $\textit{pairs} \gets \{(h_i, h_j) : h_i, h_j \in \mathcal{H}_{\mathcal{G}},\; h_i \neq h_j,\; \text{not previously analyzed}\}$
\For{each $(h_i, h_j) \in \textit{pairs}$} \Comment{Compute pairwise centroid distances}
    \State $d(h_i, h_j) \gets \|\bar{\boldsymbol{\varphi}}(h_i) - \bar{\boldsymbol{\varphi}}(h_j)\|_2$
\EndFor
\State $\textit{pairs} \gets$ top-$N_{\text{cand}}$ elements of $\textit{pairs}$ by $d$ in decreasing order \Comment{Keep most distant}
\State Sort $\textit{pairs}$ by $d$ in decreasing order; let $r$ be the rank \Comment{$r{=}1$: most distant}
\State $p_r \gets (r+1)^{-1} / \sum_{r'=1}^{N_{\text{cand}}} (r'+1)^{-1}$ \Comment{Inverse-rank probability}
\State $\mathcal{Q} \gets$ sample $N_{\text{pair}}$ pairs from $\textit{pairs}$ without replacement under $p_r$
\State \Return $\mathcal{Q}$
\end{algorithmic}
\end{algorithm}

\subsection{Instance Space Analysis and Instance Evolution}
\label{sec:pseudo_isa}

Algorithm~\ref{alg:isa} (\textsc{InstanceSpaceAnalysis}) orchestrates the two-phase sequential instance evolution for a given heuristic pair. For each phase, the instance-evolver LLM $\boldsymbol{\pi}_{\text{instance-evolver}}$ generates a pair-specific generate-and-transform function $g_{\text{inst}}$, which is then used within Algorithm~\ref{alg:evolve_inst} (\textsc{EvolveInstances}) to iteratively refine instances through selection pressure. The early-exit mechanism detects dominance: if one heuristic wins even on adversarially generated instances, the pair is classified as dominated, with that heuristic as its clear winner, without proceeding to Phase~B or decision tree training. Otherwise, a decision tree is trained on the combined instance set to identify interpretable feature-space regions of specialization.

\enlargethispage{5\baselineskip}
Algorithm~\ref{alg:evolve_inst} implements the evolutionary loop itself. The population is bootstrapped from grid-stored seed instances through the operator's generation mode; each generation then evaluates the cost-ratio margin, truncates to the top-$N_{\text{inst}}$, applies the operator's transform mode, and keeps the best of parents and offspring. A stagnation counter or a margin threshold triggers early termination.

\begin{algorithm}[H]
\caption{\textsc{InstanceSpaceAnalysis}: Instance Space Analysis for a Heuristic Pair}
\label{alg:isa}
\small
\begin{algorithmic}[1]
\Require Heuristics $h_i, h_j$, grid archive $\mathcal{G}$, instance-evolver LLM $\boldsymbol{\pi}_{\text{instance-evolver}}$, instance count $N_{\text{inst}}$, generations $G$, stagnation limit $S$, tree depth $D_{\max}$, leaf ratio $\alpha$
\Ensure Analysis result: dominated pair (clear winner) or decision tree $\mathcal{T}_{ij}$ with evolved instances and per-leaf assignments
\Statex \textbf{--- Phase A: evolve instances where $h_j$ dominates ---}
\State $X_{\text{seed}}^{A} \gets \Call{GetStoredInstances}{\mathcal{G}, h_j}$ \Comment{Instances from cells where $h_j$ is specialist}
\State $g_{\text{inst}}^{A} \gets \boldsymbol{\pi}_{\text{instance-evolver}}(h_i, h_j, \text{``hard for } h_i\text{''})$ \Comment{Eq.~\ref{eq:inst_gen_llm}}
\State $\mathcal{X}_{h_j}, m^*_{A} \gets \Call{EvolveInstances}{g_{\text{inst}}^{A}, h_i, h_j, \max, X_{\text{seed}}^{A}, N_{\text{inst}}, G, S}$
\Statex \hfill $\triangleright$ Maximize $m_{h_i,h_j}(x)$: find instances hard for $h_i$ (Alg.~\ref{alg:evolve_inst})
\If{$m^*_{A} < 1.0$} \Comment{$h_i$ outperforms $h_j$ even on adversarial instances}
    \State \Return $(\text{type}=\textit{dominated},\; \text{winner}=h_i,\; \text{instances}=\mathcal{X}_{h_j})$
\EndIf
\Statex \textbf{--- Phase B: evolve instances where $h_i$ dominates ---}
\State $X_{\text{seed}}^{B} \gets \Call{GetStoredInstances}{\mathcal{G}, h_i}$ \Comment{Instances from cells where $h_i$ is specialist}
\State $g_{\text{inst}}^{B} \gets \boldsymbol{\pi}_{\text{instance-evolver}}(h_i, h_j, \text{``hard for } h_j\text{''})$ \Comment{Eq.~\ref{eq:inst_gen_llm}}
\State $\mathcal{X}_{h_i}, m^*_{B} \gets \Call{EvolveInstances}{g_{\text{inst}}^{B}, h_i, h_j, \min, X_{\text{seed}}^{B}, N_{\text{inst}}, G, S}$
\Statex \hfill $\triangleright$ Minimize $m_{h_i,h_j}(x)$: find instances hard for $h_j$ (Alg.~\ref{alg:evolve_inst})
\If{$m^*_{B} > 1.0$} \Comment{$h_j$ outperforms $h_i$ even on adversarial instances}
    \State \Return $(\text{type}=\textit{dominated},\; \text{winner}=h_j,\; \text{instances}=\mathcal{X}_{h_i})$
\EndIf
\Statex \textbf{--- Phase C: decision-tree training ---}
\State $X_{\text{all}} \gets \mathcal{X}_{h_j} \cup \mathcal{X}_{h_i}$ \Comment{Combine instances from both phases}
\For{each $x \in X_{\text{all}}$}
    \State $\boldsymbol{\varphi}(x) \gets \Call{ExtractFeatures}{x}$ \Comment{Compute instance features}
    \State $y(x) \gets \mathbb{I}[P(h_j, x) > P(h_i, x)]$ \Comment{Label: $1$ if $h_j$ wins}
\EndFor
\State $\mathcal{T}_{ij} \gets \Call{TrainDecisionTree}{\{(\boldsymbol{\varphi}(x), y(x))\}_{x \in X_{\text{all}}}, D_{\max}, \lfloor \alpha \cdot |X_{\text{all}}| \rfloor}$
\State $\mathcal{R}_{h_i}, \mathcal{R}_{h_j} \gets \Call{ExtractRules}{\mathcal{T}_{ij}}$ \Comment{Per-class rule sets with leaf assignments}
\State \Return $(\text{type}=\textit{discriminated},\; \text{tree}=\mathcal{T}_{ij},\; \text{instances}=X_{\text{all}},\; \text{rules}=(\mathcal{R}_{h_i}, \mathcal{R}_{h_j}))$
\end{algorithmic}
\end{algorithm}

\begin{algorithm}[H]
\caption{\textsc{EvolveInstances}: LLM-Guided Evolutionary Instance Generation}
\label{alg:evolve_inst}
\small
\begin{algorithmic}[1]
\Require Generate-and-transform operator $g_{\text{inst}}$, heuristics $h_i, h_j$, direction $\in \{\max, \min\}$, seed instances $X_{\text{seed}}$, instances per generation $N_{\text{inst}}$, generations $G$, stagnation limit $S$
\Ensure Evolved instance set $\mathcal{X}$, best margin $m^*$
\State $X \gets g_{\text{inst}}(X_{\text{seed}}, N_{\text{inst}})$ \Comment{Bootstrap: generation mode grows the seeds to $N_{\text{inst}}$ instances}
\State $m^* \gets$ worst possible;\; $\textit{stagnation} \gets 0$
\For{$g_{\text{idx}} = 1, \ldots, G$}
    \For{each $x \in X$} \Comment{Evaluate cost-ratio margin}
        \State $m(x) \gets f(h_i(x))/f(h_j(x))$ \Comment{Eq.~\ref{eq:margin}}
    \EndFor
    \If{best $m(x)$ in $X$ improves $m^*$ w.r.t.\ direction} \Comment{Track best margin for dominance detection}
        \State $m^* \gets$ best $m(x)$;\; $\textit{stagnation} \gets 0$
    \Else
        \State $\textit{stagnation} \gets \textit{stagnation} + 1$
    \EndIf
    \If{$\textit{stagnation} \geq S$ \textbf{or} $m^*$ exceeds the margin threshold} \Comment{Early termination}
        \State \textbf{break}
    \EndIf
    \State $X \gets$ top $N_{\text{inst}}$ instances by $m(x)$ \Comment{Decreasing order if direction $= \max$, else increasing}
    \State $X' \gets g_{\text{inst}}(X, N_{\text{inst}})$ \Comment{Transform mode: incremental structural modifications}
    \State $X \gets$ top $N_{\text{inst}}$ instances of $X \cup X'$ by $m(x)$ \Comment{$(\mu{+}\lambda)$ truncation over parents and offspring}
\EndFor
\State \Return $X, m^*$
\end{algorithmic}
\end{algorithm}

\subsection{Grid Update and Insight Generation}
\label{sec:pseudo_grid_update}

\enlargethispage{3\baselineskip}
Algorithm~\ref{alg:grid_update} (\textsc{UpdateGridAndGenerateInsights}) is the central mechanism that connects instance evolution to heuristic specialization. It handles two cases. In the \emph{dominated} case, the clear winner directly replaces the loser's cells and evolved instances expand the grid into new regions. In the \emph{discriminated} case, the algorithm processes each decision-tree leaf: only validated instances (where the predicted heuristic actually performs better) are assigned to grid cells. Cells at full capacity dynamically replace their worst-performing instance if the new one better represents the region. The reflection LLM then generates three contrastive insights, which are stored in the grid cells of validated instances, tagged with the current iteration for recency-based retrieval.

\begin{algorithm}[H]
\caption{\textsc{UpdateGridAndGenerateInsights}: Grid Update and Insight Generation}
\label{alg:grid_update}
\small
\begin{algorithmic}[1]
\Require Grid $\mathcal{G}$, heuristics $h_i, h_j$, ISA result $\textit{analysis}_{ij}$, reflection LLM $\boldsymbol{\pi}_{\text{reflection}}$, max instances per cell $M$, iteration $t$
\Ensure Updated grid $\mathcal{G}$, crossover insight $I_{\times}$
\If{$\textit{analysis}_{ij}.\text{type}$ = \textbf{dominated}}
    \State $h_w \gets \textit{analysis}_{ij}.\text{winner}$;\; $h_l \gets$ the other heuristic in $(h_i, h_j)$
    \Statex \quad\textbf{--- Replace loser's cells with winner ---}
    \State $\mathcal{C}_l \gets \{\mathbf{k} : h_{\mathbf{k}} = h_l\}$ \Comment{Cells where loser is specialist}
    \For{each $\mathbf{k} \in \mathcal{C}_l$} \Comment{Directly replace loser with winner}
        \State $h_{\mathbf{k}} \gets h_w$;\; $\nu_{\mathbf{k}} \gets \nu_{\mathbf{k}} + 1$
    \EndFor
    \Statex \quad\textbf{--- Assign evolved instances to new cells ---}
    \For{each $x \in \textit{analysis}_{ij}.\text{instances}$} \Comment{Use evolved instances to expand grid}
        \State $\mathbf{k} \gets \kappa(\boldsymbol{\varphi}(x))$
        \If{$\mathbf{c}_{\mathbf{k}} \notin \mathcal{G}$} \Comment{Only create new cells}
            \State $\mathcal{G}.\text{add}(\mathbf{c}_{\mathbf{k}})$;\; $X_{\mathbf{k}} \gets \{x\}$;\; $h_{\mathbf{k}} \gets h_w$
        \EndIf
    \EndFor
    \Statex \quad\textbf{--- Generate and store global reflection ---}
    \State $I_{\times} \gets \boldsymbol{\pi}_{\text{reflection}}(h_w, h_l, \text{``global''})$ \Comment{Global reflection as crossover insight}
    \For{each $\mathbf{k} \in \mathcal{C}(h_w) \cup \mathcal{C}_l$}
        \State $\Xi_{\mathbf{k}}.\text{add}(\text{insight}=I_{\times},\; \text{iteration}=t)$
    \EndFor
\ElsIf{$\textit{analysis}_{ij}.\text{type}$ = \textbf{discriminated}}
    \State $\mathcal{T}_{ij} \gets \textit{analysis}_{ij}.\text{tree}$;\; $\mathcal{R}_{h_i}, \mathcal{R}_{h_j} \gets \textit{analysis}_{ij}.\text{rules}$
    \Statex \quad\textbf{--- Assign validated instances to grid per leaf ---}
    \For{each leaf $\ell$ in $\mathcal{T}_{ij}$} \Comment{Each leaf predicts a winning heuristic}
        \State $h_w \gets h_i$ if $\ell$ predicts class $i$, else $h_j$;\; $h_o \gets h_j$ if $h_w = h_i$, else $h_i$
        \For{each $x$ in leaf $\ell$ \textbf{where} $P(h_w, x) > P(h_o, x)$} \Comment{Only validated instances}
            \State $\mathbf{k} \gets \kappa(\boldsymbol{\varphi}(x))$ \Comment{Map to grid cell (Eq.~\ref{eq:discretization})}
            \If{$\mathbf{c}_{\mathbf{k}} \notin \mathcal{G}$} \Comment{New cell: create and assign instance + heuristic}
                \State $\mathcal{G}.\text{add}(\mathbf{c}_{\mathbf{k}})$;\; $X_{\mathbf{k}} \gets \{x\}$;\; $h_{\mathbf{k}} \gets h_w$
            \ElsIf{$|X_{\mathbf{k}}| < M$} \Comment{Cell has capacity: add instance}
                \State $X_{\mathbf{k}} \gets X_{\mathbf{k}} \cup \{x\}$
            \Else \Comment{Cell full: dynamically update stored instances}
                \State $x_{\text{worst}} \gets \arg\min_{x' \in X_{\mathbf{k}}} P(h_{\mathbf{k}}, x')$ \Comment{Worst stored instance under current specialist}
                \If{$P(h_w, x) > P(h_{\mathbf{k}}, x_{\text{worst}})$} \Comment{New instance better represents this region}
                    \State $X_{\mathbf{k}} \gets (X_{\mathbf{k}} \setminus \{x_{\text{worst}}\}) \cup \{x\}$
                \EndIf
            \EndIf
            \If{$h_{\mathbf{k}} \neq h_w$ \textbf{and} $P(h_w; X_{\mathbf{k}}) > P(h_{\mathbf{k}}; X_{\mathbf{k}})$} \Comment{Replace heuristic if better}
                \State $h_{\mathbf{k}} \gets h_w$;\; $\nu_{\mathbf{k}} \gets \nu_{\mathbf{k}} + 1$
            \EndIf
        \EndFor
    \EndFor
    \Statex \quad\textbf{--- Generate and store contrastive insights ---}
    \State $\{I_{h_i}, I_{h_j}, I_{\times}\} \gets \boldsymbol{\pi}_{\text{reflection}}(h_i, h_j)$ \Comment{Three mechanism-level insights (Eq.~\ref{eq:reflection})}
    \For{each leaf $\ell$ in $\mathcal{T}_{ij}$} \Comment{Store leaf's insight in validated cells}
        \State $h_w \gets h_i$ if $\ell$ predicts class $i$, else $h_j$;\; $h_o \gets h_j$ if $h_w = h_i$, else $h_i$
        \State $V_\ell \gets \{x \in \ell : P(h_w, x) > P(h_o, x)\}$ \Comment{Instances where prediction is correct}
        \State $I_\ell \gets I_{h_i}$ if $h_w = h_i$, else $I_{h_j}$ \Comment{Select matching insight}
        \For{each unique $\mathbf{k} \in \{\kappa(\boldsymbol{\varphi}(x)) : x \in V_\ell\}$}
            \State $\Xi_{\mathbf{k}}.\text{add}(\text{insight}=I_\ell,\; \text{iteration}=t)$
        \EndFor
    \EndFor
\EndIf
\State \Return $\mathcal{G}, I_{\times}$ \Comment{Updated grid and crossover insight}
\end{algorithmic}
\end{algorithm}

\subsection{Crossover, Mutation, and Local Evaluation}
\label{sec:pseudo_crossover_mutation}

Algorithm~\ref{alg:crossover} (\textsc{LLMCrossover}) receives the crossover insight $I_{\times}$ produced by Algorithm~\ref{alg:grid_update} and prompts the crossover LLM to generate an offspring that combines the complementary strengths of both parents. The offspring is evaluated on a cube centered at the midpoint of the parents' feature-space centroids.

Algorithm~\ref{alg:mutation} (\textsc{InsightDrivenMutation}) performs region-aware mutation in three steps. First, a parent heuristic is sampled with probability proportional to its grid coverage (broadly successful heuristics are more likely to produce competitive variants). Second, a target cube is selected by staleness-weighted sampling, directing mutations toward under-explored regions. Third, localized insights are retrieved from the target cube via Algorithm~\ref{alg:insights} (\textsc{RetrieveInsights}), which prioritizes current-iteration insights by novelty (rarest first), then fills remaining slots with older insights sorted by recency and then novelty.

Algorithm~\ref{alg:eval_cube} (\textsc{EvaluateOnCube}) implements the cell-level replacement rule used by both crossover and mutation: the candidate replaces the incumbent in each cube cell where it achieves strictly higher average performance, and each evaluated candidate advances the budget counter $e$ of Algorithm~\ref{alg:main}.

\begin{algorithm}[H]
\caption{\textsc{LLMCrossover}: LLM-Guided Crossover}
\label{alg:crossover}
\small
\begin{algorithmic}[1]
\Require Crossover LLM $\boldsymbol{\pi}_{\text{crossover}}$, heuristics $h_i, h_j$, crossover insight $I_{\times}$, cube axis fraction $\rho$, min cells $T$
\Ensure Offspring heuristic $h'$
\State $h' \gets \boldsymbol{\pi}_{\text{crossover}}(h_i, h_j, I_{\times})$ \Comment{Generate offspring guided by crossover insight (Eq.~\ref{eq:crossover})}
\Statex \textbf{--- Evaluate offspring on parent-centered cube ---}
\State $\bar{\mathbf{k}} \gets \lfloor (\bar{\boldsymbol{\varphi}}(h_i) + \bar{\boldsymbol{\varphi}}(h_j))/2 \rceil$ \Comment{Midpoint of parent centroids}
\State $\mathcal{B} \gets \Call{BuildCube}{\bar{\mathbf{k}}, \rho, T}$ \Comment{Cube with $\geq T$ filled cells}
\State $\Call{EvaluateOnCube}{\mathcal{G}, h', \mathcal{B}}$ \Comment{Replace if better (Alg.~\ref{alg:eval_cube})}
\State \Return $h'$
\end{algorithmic}
\end{algorithm}

\begin{algorithm}[H]
\caption{\textsc{InsightDrivenMutation}: Region-Aware Insight-Driven Mutation}
\label{alg:mutation}
\small
\begin{algorithmic}[1]
\Require Grid archive $\mathcal{G}$, mutation LLM $\boldsymbol{\pi}_{\text{mutation}}$, truncation size $N_{\text{cand}}$, $C$ candidate cubes, max insights $L$, cube axis fraction $\rho$, min cells $T$, iteration $t$
\Ensure Mutated heuristic $h''$, target cube $\mathcal{B}$
\Statex \textbf{--- Step 1: coverage-ranked parent selection ---}
\State $\mathcal{H}_{\mathcal{G}} \gets \{h : \exists\, \mathbf{k} \text{ s.t.\ } h_{\mathbf{k}} = h\}$ \Comment{Collect unique heuristics in $\mathcal{G}$}
\If{$|\mathcal{H}_{\mathcal{G}}| > N_{\text{cand}}$} \Comment{Keep only the $N_{\text{cand}}$ heuristics with highest coverage $n(h)$}
    \State $\mathcal{H}_{\mathcal{G}} \gets$ the $N_{\text{cand}}$ elements of $\mathcal{H}_{\mathcal{G}}$ with largest $n(h)$
\EndIf
\State Sort $\mathcal{H}_{\mathcal{G}}$ by $n(h)$ in decreasing order; let $r(h)$ be the rank \Comment{Highest coverage ranked first}
\State $p(h) \gets (r(h)+1)^{-1} / \sum_{h' \in \mathcal{H}_{\mathcal{G}}} (r(h')+1)^{-1}$ \Comment{Inverse-rank probability}
\State $h_{\text{parent}} \sim p(\cdot)$ \Comment{Sample parent biased toward high-coverage heuristics}
\Statex \textbf{--- Step 2: staleness-guided target region selection ---}
\For{$c = 1, \ldots, C$} \Comment{Generate $C$ candidate cubes centered at random filled cells}
    \State $\mathbf{k}_c \sim \text{Uniform}(\{\mathbf{k} : h_{\mathbf{k}} \neq \varnothing\})$
    \State $\mathcal{B}_c \gets \Call{BuildCube}{\mathbf{k}_c, \rho, T}$ \Comment{Cube with $\geq T$ filled cells}
    \State $\bar{\sigma}_c \gets \frac{1}{|\mathcal{B}_c|} \sum_{\mathbf{c}_{\mathbf{k}} \in \mathcal{B}_c} (\nu_{\mathbf{k}} + 1)^{-1}$ \Comment{Average staleness}
\EndFor
\State Sort $\{\mathcal{B}_c\}$ by $\bar{\sigma}_c$ in decreasing order; let $r(c)$ be the rank \Comment{Stalest cubes ranked first}
\State $p(c) \gets (r(c)+1)^{-1} / \sum_{c'=1}^{C} (r(c')+1)^{-1}$ \Comment{Inverse-rank probability}
\State $\mathcal{B} \gets \mathcal{B}_{\tilde{c}}$ where $\tilde{c} \sim p(\cdot)$ \Comment{Sample cube biased toward stale regions}
\Statex \textbf{--- Step 3: insight retrieval and LLM mutation ---}
\State $\hat{\Xi} \gets \Call{RetrieveInsights}{\mathcal{G}, \mathcal{B}, L, t}$ \Comment{Up to $L$ prioritized insights from cube (Alg.~\ref{alg:insights})}
\State $h'' \gets \boldsymbol{\pi}_{\text{mutation}}(h_{\text{parent}}, \mathrm{bundle}(\hat{\Xi}))$ \Comment{Generate mutant heuristic (Eq.~\ref{eq:mutation})}
\State \Return $h'', \mathcal{B}$
\end{algorithmic}
\end{algorithm}

\begin{algorithm}[H]
\caption{\textsc{RetrieveInsights}: Prioritized Insight Retrieval from Grid Cells}
\label{alg:insights}
\small
\begin{algorithmic}[1]
\Require Grid archive $\mathcal{G}$, cube cells $\mathcal{B}$, max insights $L$, current iteration $t$
\Ensure Ordered list of up to $L$ unique insights
\Statex \Comment{Each entry $s \in \Xi_{\mathbf{k}}$ has two fields: $s.\text{insight}$ and $s.\text{iteration}$}
\State $\hat{\Xi} \gets [\,]$;\; $\textit{seen} \gets \emptyset$ \Comment{Output list; set for deduplication}
\Statex \textbf{--- Priority 1: current-iteration insights ---}
\State $\textit{coverage}(\cdot) \gets 0$ \Comment{Count how many cube cells share each insight}
\For{each $\mathbf{c}_{\mathbf{k}} \in \mathcal{B}$}
    \For{each $s \in \Xi_{\mathbf{k}}$ \textbf{where} $s.\text{iteration} = t$} \Comment{Only insights from current iteration}
        \State $\textit{coverage}(s.\text{insight}) \gets \textit{coverage}(s.\text{insight}) + 1$
    \EndFor
\EndFor
\For{each $s.\text{insight}$ in increasing order of $\textit{coverage}$} \Comment{Novelty-first: rarest insights first}
    \If{$s.\text{insight} \notin \textit{seen}$ \textbf{and} $|\hat{\Xi}| < L$}
        \State $\hat{\Xi}.\text{append}(s.\text{insight})$;\; $\textit{seen} \gets \textit{seen} \cup \{s.\text{insight}\}$
    \EndIf
\EndFor
\Statex \textbf{--- Priority 2: previous-iteration insights ---}
\If{$|\hat{\Xi}| < L$} \Comment{Fill remaining slots with older insights}
    \State $\textit{older} \gets \{s \in \Xi_{\mathbf{k}} : \mathbf{c}_{\mathbf{k}} \in \mathcal{B},\; s.\text{iteration} < t\}$
    \For{each $s \in \textit{older}$ in decreasing order of $s.\text{iteration}$, then increasing order of $\textit{coverage}(s.\text{insight})$}
        \If{$s.\text{insight} \notin \textit{seen}$ \textbf{and} $|\hat{\Xi}| < L$} \Comment{Most recent, then rarest}
            \State $\hat{\Xi}.\text{append}(s.\text{insight})$;\; $\textit{seen} \gets \textit{seen} \cup \{s.\text{insight}\}$
        \EndIf
    \EndFor
\EndIf
\State \Return $\hat{\Xi}$
\end{algorithmic}
\end{algorithm}

\begin{algorithm}[H]
\caption{\textsc{EvaluateOnCube}: Local Heuristic Evaluation and Grid Update}
\label{alg:eval_cube}
\small
\begin{algorithmic}[1]
\Require Grid archive $\mathcal{G}$, candidate heuristic $h'$, cube $\mathcal{B}$, budget counter $e$
\For{each cell $\mathbf{c}_{\mathbf{k}} \in \mathcal{B}$} \Comment{Evaluate candidate on each cell in the cube}
    \State $P(h'; X_{\mathbf{k}}) \gets \frac{1}{|X_{\mathbf{k}}|} \sum_{x \in X_{\mathbf{k}}} P(h', x)$ \Comment{Avg.\ performance on cell's instances}
    \If{$P(h'; X_{\mathbf{k}}) > P(h_{\mathbf{k}}; X_{\mathbf{k}})$} \Comment{Replace specialist if $h'$ is better}
        \State $h_{\mathbf{k}} \gets h'$;\; $\nu_{\mathbf{k}} \gets \nu_{\mathbf{k}} + 1$
    \EndIf
\EndFor
\State $e \gets e + 1$ \Comment{Advance the budget counter of Alg.~\ref{alg:main}}
\end{algorithmic}
\end{algorithm}

\subsection{Test-Time Portfolio Extraction}
\label{sec:pseudo_portfolio}

Algorithm~\ref{alg:portfolio} (\textsc{GreedyPortfolio}) implements the greedy complementary selection of Section~\ref{sec:method_select}. It maintains the per-instance best-of-set cost $b(x)$ over $\mathcal{D}_{\text{val}}$ and, at each step, adds the heuristic that most reduces its mean; a single run yields nested Top-$k$ portfolios for every $k$.

\begin{algorithm}[H]
\caption{\textsc{GreedyPortfolio}: Complementary Portfolio Extraction}
\label{alg:portfolio}
\small
\begin{algorithmic}[1]
\Require Archive heuristics $\mathcal{H}_{\mathcal{G}}$, validation instances $\mathcal{D}_{\text{val}}$
\Ensure Ordered portfolio $A$ (its length-$k$ prefix is the Top-$k$ portfolio $A_k$)
\State $A \gets [\,]$;\; $b(x) \gets \infty$ for all $x \in \mathcal{D}_{\text{val}}$ \Comment{Best-of-set cost so far}
\For{$j = 1, \ldots, |\mathcal{H}_{\mathcal{G}}|$}
    \State $h^{(j)} \gets \operatorname*{arg\,min}_{h \in \mathcal{H}_{\mathcal{G}} \setminus A} \frac{1}{|\mathcal{D}_{\text{val}}|} \sum_{x} \min\big(b(x),\, f(h(x))\big)$ \Comment{Eq.~\ref{eq:portfolio}}
    \State $A.\text{append}(h^{(j)})$
    \State $b(x) \gets \min\big(b(x),\, f(h^{(j)}(x))\big)$ for all $x \in \mathcal{D}_{\text{val}}$
\EndFor
\State \Return $A$
\end{algorithmic}
\end{algorithm}

\section{Prompt Templates}
\label{app:prompts}

The prompts of the five LLM roles are provided below: initialization, the instance evolver ($\boldsymbol{\pi}_{\text{instance-evolver}}$), contrastive reflection ($\boldsymbol{\pi}_{\text{reflection}}$), crossover ($\boldsymbol{\pi}_{\text{crossover}}$), and mutation ($\boldsymbol{\pi}_{\text{mutation}}$). Each role pairs a fixed system prompt with a user prompt populated from the current search state; fields in curly braces are filled at query time. The problem descriptions, function descriptions, and function signatures follow the formats used in ReEvo~\citep{ye2024reevo}. Problem-specific and query-dependent passages are abbreviated inside the boxes as brief \texttt{\#} comments. The crossover and mutation roles return the generated heuristic as a Python function; for $\boldsymbol{\pi}_{\text{instance-evolver}}$ and $\boldsymbol{\pi}_{\text{reflection}}$, a representative response follows each prompt: the pair-specific generate-and-transform operator $g_{\text{inst}}$ (Eq.~\ref{eq:inst_gen_llm}) and the three contrastive insights (Eq.~\ref{eq:reflection}).

\begin{tcolorbox}[
    colback=blue!6,
    colframe=blue!55!black,
    title=Initialization Prompt,
    fonttitle=\bfseries,
    breakable,
    enhanced,
    arc=2mm,
]
\scriptsize
\begin{verbatim}
[SYSTEM PROMPT]
You are an expert in the domain of optimization heuristics. Your task is to design heuristics
that can effectively solve optimization problems.
Your response outputs Python code and nothing else. Format your code as a Python code string:
"```python ... ```".

------------------------------------------------------------

[USER PROMPT]
Write a {func_name} function for {problem_desc}
{func_desc}

Function signature:
{func_signature}

# brief diversity directive: design ONE distinctive selection rule that is strong on some
# structural family of instances rather than mediocre everywhere

Requirements:
# brief functional requirements: validity of the returned decision, robustness to edge cases,
# and an efficiency guard on per-call computation

Format your response as:
```python
[your generated code here]
```
\end{verbatim}
\end{tcolorbox}

\begin{tcolorbox}[
    colback=green!8,
    colframe=green!45!black,
    title={Instance-Evolver Prompt ($\boldsymbol{\pi}_{\text{instance-evolver}}$)},
    fonttitle=\bfseries,
    breakable,
    enhanced,
    arc=2mm,
]
\scriptsize
\begin{verbatim}
[SYSTEM PROMPT]
You are an expert in combinatorial optimization and instance generation. Your task is to design
Python functions that generate problem instances with different structural properties.
Your response outputs Python code and nothing else. Format your code as a Python code string:
"```python ... ```".

------------------------------------------------------------

[USER PROMPT]
{problem_context}   # short problem description: instance encoding and objective

You are given two constructive heuristic solvers for the {problem_desc}:

[Heuristic No.1]
{heuristic_1_code}

[Heuristic No.2]
{heuristic_2_code}

[Task]
{generation_direction}
  # one of two symmetric directions depending on the targeted heuristic of the pair: reshape
  # the instance structure to make instances HARDER for one heuristic and EASIER for the other

You must write a function that can both GENERATE new instances and TRANSFORM existing ones.
This function will be called iteratively in an evolutionary loop:
- First call: receives a small number of seed instances (fewer than n_instances). You must
  generate additional instances to reach n_instances total.
- Subsequent calls: receives n_instances instances. Apply incremental transformations to make
  them harder for the target heuristic.

[Function Signature]
# problem-dependent; for TSP:
def generate_and_transform_instances(instances: list, n_cities: int, n_instances: int) -> list:
    """
    Generate new instances or transform existing ones to produce harder variants.

    This function operates in two modes based on the input:
    - If len(instances) < n_instances: Generate new instances (using the seeds as structural
      reference) to reach n_instances total.
    - If len(instances) >= n_instances: Transform existing instances with incremental
      modifications.

    In both modes, the returned instances should exploit the target heuristic's weaknesses.

    Args:
        instances: List of numpy arrays, each with shape (n_cities, 2), coordinates in
                   [0, 1]^2. May contain fewer instances than n_instances (seed mode) or
                   equal/more (transform mode).
        n_cities: Number of cities per instance.
        n_instances: Target number of instances to return.

    Returns:
        List of exactly n_instances numpy arrays, each with shape (n_cities, 2),
        coordinates in [0, 1]^2.
    """

[How to Approach This]
1. Read both heuristics and identify their key decision-making differences.
2. At the start of your function, check `if len(instances) < n_instances:` to decide the mode.
3. GENERATION MODE (len(instances) < n_instances):
   - You need to produce `n_instances - len(instances)` new instances.
   - For each new instance, start from a copy of a random seed (pick from `instances` with
     replacement), then apply structural perturbations: move nodes to form clusters, create
     corridors, shift density.
   - Combine the original seeds with the newly generated instances.
4. TRANSFORM MODE (len(instances) >= n_instances):
   - Apply incremental modifications to each instance: move nodes to form/break clusters,
     create corridors, adjust density.
   - Keep transformations incremental -- small changes per generation.
5. Return exactly n_instances instances in both modes.

[Code Requirements]
- Use only `numpy` (imported as `np`) and `math`. Use `rng = np.random.default_rng()` for
  randomness.
- CRITICAL: Each returned instance must have exactly (n_cities, 2) shape. Never add or remove
  nodes.
- CRITICAL: Return exactly n_instances instances.
- Do NOT use deprecated `.ptp()`. Use `arr.max() - arr.min()` instead.
- Ensure `rng.integers(low, high)` has `low < high`. Use `float(x)` or `int(x)` for scalar
  conversion.
- When indexing numpy results, ensure the result is an array, not a scalar. Use
  `np.atleast_1d()` if needed.
- Clip all coordinates to [0, 1] before returning.
- Ensure all parentheses and brackets are matched.
- Keep the code simple -- straightforward loops over instances are safer than complex
  vectorized operations.

Output ONLY the function code enclosed in a Python code block: ```python ... ```.
\end{verbatim}
\end{tcolorbox}

\begin{tcolorbox}[
    colback=green!8,
    colframe=green!45!black,
    title={Instance-Evolver Example Response ($g_{\text{inst}} = \boldsymbol{\pi}_{\text{instance-evolver}}(h_i, h_j, \text{direction})$)},
    fonttitle=\bfseries,
    breakable,
    enhanced,
    arc=2mm,
]
\scriptsize
\begin{verbatim}
def generate_and_transform_instances(instances: list, n_cities: int, n_instances: int) -> list:
    """
    Generate new TSP instances or transform existing ones to produce harder variants.
    - Generation mode when len(instances) < n_instances: create new instances by perturbing seeds.
    - Transformation mode when len(instances) >= n_instances: apply incremental perturbations.
    All outputs maintain exactly (n_cities, 2) shape with coordinates in [0, 1].
    """
    import numpy as np
    import math

    rng = np.random.default_rng()

    # Initialize a persistent rotation phase to rotate the structural layout across calls
    if not hasattr(generate_and_transform_instances, "_rot_phase"):
        generate_and_transform_instances._rot_phase = rng.integers(0, 1 << 30) / (1 << 30) * 2 * math.pi

    def transform_points(points, theta, jitter_scale=0.003, warp_ampl=0.04):
        """Rotate around center, apply a mild warp and small jitter, then clip to [0,1]."""
        pts = np.asarray(points, dtype=float).copy()
        c = 0.5
        dx = pts[:, 0] - c
        dy = pts[:, 1] - c
        cth = math.cos(theta)
        sth = math.sin(theta)
        x = c + cth * dx - sth * dy
        y = c + sth * dx + cth * dy
        # Mild corridor-like warp to induce structured dispersion without clustering
        x = x + warp_ampl * np.sin(2 * np.pi * x)
        y = y + warp_ampl * np.cos(2 * np.pi * y)
        if jitter_scale > 0:
            jitter = rng.normal(scale=jitter_scale, size=pts.shape)
            x += jitter[:, 0]
            y += jitter[:, 1]
        x = np.clip(x, 0.0, 1.0)
        y = np.clip(y, 0.0, 1.0)
        pts[:, 0] = x
        pts[:, 1] = y
        return pts

    if len(instances) < n_instances:
        # GENERATION MODE
        out = []
        # Copy existing seeds first to preserve them
        for inst in instances:
            out.append(np.asarray(inst, dtype=float).copy())
        seeds = [np.asarray(inst, dtype=float).copy() for inst in instances] if len(instances) > 0 else []
        needed = n_instances - len(instances)

        for _ in range(needed):
            if seeds:
                base = seeds[rng.integers(0, len(seeds))]
                base_pt = base.copy()
            else:
                base_pt = rng.random((n_cities, 2))

            theta = generate_and_transform_instances._rot_phase
            # Progress phase to rotate structure across generations
            generate_and_transform_instances._rot_phase = (theta + math.pi / 6) % (2 * math.pi)  # 30 degrees
            transformed = transform_points(base_pt, theta, jitter_scale=0.006, warp_ampl=0.06)
            out.append(transformed)

        # Ensure exactly n_instances
        return out[:n_instances]
    else:
        # TRANSFORM MODE
        out = []
        for idx in range(n_instances):
            inst = instances[idx]
            pts = np.asarray(inst, dtype=float).copy()
            theta = generate_and_transform_instances._rot_phase
            generate_and_transform_instances._rot_phase = (theta + math.pi / 12) % (2 * math.pi)  # 15 degrees
            transformed = transform_points(pts, theta, jitter_scale=0.003, warp_ampl=0.04)
            out.append(transformed)
        return out
\end{verbatim}
\end{tcolorbox}

\begin{tcolorbox}[
    colback=orange!10,
    colframe=orange!75!black,
    title={Contrastive Reflection Prompt ($\boldsymbol{\pi}_{\text{reflection}}$)},
    fonttitle=\bfseries,
    breakable,
    enhanced,
    arc=2mm,
]
\scriptsize
\begin{verbatim}
[SYSTEM PROMPT]
You are an expert in constructive heuristics for {problem_desc}.
You are shown two {func_name} heuristics that excel on DIFFERENT kinds of instances.
Name the distinct algorithmic MECHANISM behind each one and turn it into a general, reusable
improvement idea.

CRITICAL -- how these insights are used later:
Each insight is STORED and later handed to a DIFFERENT heuristic to guide its mutation. That
consumer does NOT see this code, this pair, or any instance features. An insight is only
useful if it is SELF-CONTAINED and ALGORITHMIC:
# brief instructions: state what the mechanism computes from the heuristic's inputs and why it
# improves the objective; no variable names or pair references; name a concrete technique,
# preferring a mechanism the code does not already use

Return exactly three insights in strict JSON (no prose, no markdown):
{
  "Insight_H1": "<35-55 words: the first heuristic's distinct mechanism stated generally, why
                 it improves the objective, and ONE concrete way to strengthen it.>",
  "Insight_H2": "<35-55 words: same for the second heuristic's distinct mechanism.>",
  "Insight_Crossover": "<35-55 words: how to combine the two mechanisms into ONE scoring
                 formula -- an additive blend of terms, NOT an if/else switch.>"
}

Be concrete and implementable, never generic ("be adaptive"). Think privately; output only
the JSON.
# appended diversity directive: if both parents share the dominant mechanism family, propose
# a structurally different mechanism instead of another re-weighting of the same terms

------------------------------------------------------------

[USER PROMPT]
Below are two {func_name} functions for {problem_desc}.
{func_desc}

[Heuristic No.1]
{heuristic_1_code}

[Heuristic No.2]
{heuristic_2_code}

Task: The two heuristics are known to win on different kinds of instances. Identify each
heuristic's distinct, reusable MECHANISM and one concrete improvement for it, then a crossover
idea that blends BOTH mechanisms into a single additive scoring formula. Follow the System
Prompt exactly: self-contained, algorithmic, no feature names, no variable names. Return the
three JSON insights.
# for a dominated pair, a single worse/better reflection contrasting the clear winner and the
# loser replaces the three insights and is stored as the pair's global crossover insight
\end{verbatim}
\end{tcolorbox}

\begin{tcolorbox}[
    colback=orange!10,
    colframe=orange!75!black,
    title={Contrastive Reflection Example Response ($\{I_{h_i}, I_{h_j}, I_{\times}\} = \boldsymbol{\pi}_{\text{reflection}}(h_i, h_j)$)},
    fonttitle=\bfseries,
    breakable,
    enhanced,
    arc=2mm,
]
\scriptsize
\begin{verbatim}
{
  "Insight_H1": "Distinct mechanism: adaptive dispersion-aware scoring that biases edges to keep
    the remaining unvisited nodes compact, penalizing moves that stretch the future set away
    from its center. This reduces detours by preserving locality, with a lightweight two-step
    lookahead to prune weak options. Strengthen by computing dispersion along two principal
    axes instead of a single centroid.",
  "Insight_H2": "Distinct mechanism: regret-guided bounded lookahead using a global lower bound
    on remaining unvisited nodes and dispersion-based shortlisting. For each candidate it
    compares bound_before and bound_after to form a regret term, biasing choices toward larger
    future savings. Improvement: switch MST bound to a tighter 1-tree lower bound for
    remaining nodes.",
  "Insight_Crossover": "Blended scoring: compute two additive terms per candidate: a local,
    dispersion-aware travel component with a shallow lookahead, and a global regret component
    from bound_before minus bound_after. Combine them with fixed weights into a single score,
    and pick the candidate with the smallest blended value."
}
\end{verbatim}
\end{tcolorbox}

\begin{tcolorbox}[
    colback=violet!8,
    colframe=violet!70!black,
    title={Crossover Prompt ($\boldsymbol{\pi}_{\text{crossover}}$)},
    fonttitle=\bfseries,
    breakable,
    enhanced,
    arc=2mm,
]
\scriptsize
\begin{verbatim}
[SYSTEM PROMPT]
Same as the initialization system prompt.

------------------------------------------------------------

[USER PROMPT]
Write a {func_name} function for {problem_desc}
{func_desc}

You are combining two parent heuristics into a stronger child. The reflection names each
parent's complementary mechanism.

[Heuristic No.1]
{heuristic_1_code}

[Heuristic No.2]
{heuristic_2_code}

[Reflection]
{reflection}   # the pair's crossover insight; for a dominated pair, its global reflection

Combine both parents' mechanisms into ONE scoring formula (an additive blend of terms, not an
if/else that switches between the parents).
# brief validity directive: the child must remain a valid decision step for the objective

[Improved code]
Write an improved function `{func_name}_v2` that blends both mechanisms.
# brief functional requirements: robustness to edge cases and an efficiency guard on per-call
# computation
Output code only, enclosed in a Python code block: ```python ... ```.
\end{verbatim}
\end{tcolorbox}

\begin{tcolorbox}[
    colback=red!6,
    colframe=red!60!black,
    title={Mutation Prompt ($\boldsymbol{\pi}_{\text{mutation}}$)},
    fonttitle=\bfseries,
    breakable,
    enhanced,
    arc=2mm,
]
\scriptsize
\begin{verbatim}
[SYSTEM PROMPT]
Same as the initialization system prompt.

------------------------------------------------------------

[USER PROMPT]
Write a {func_name} function for {problem_desc}
{func_desc}

You are improving ONE heuristic. Below is an algorithmic insight (a general mechanism that
tends to improve the objective) and the current code.

[Insight]
{reflection}   # the bundle of up to L=2 insights retrieved from the target cube

[Code to mutate]
{elitist_code}

[How to mutate]
- Produce a GENUINELY modified heuristic: add or strengthen the mechanism in the insight, or
  introduce a clearly different mechanism if it looks more promising. Do NOT return the input
  essentially unchanged, and do NOT merely tweak a constant.
# brief validity directive: the mutant must remain a valid decision step for the objective

[Improved code]
Write a mutated function `{func_name}_v2`.
# brief functional requirements: robustness to edge cases and an efficiency guard on per-call
# computation
Output code only, enclosed in a Python code block: ```python ... ```.
\end{verbatim}
\end{tcolorbox}

\end{document}